\documentclass[10pt]{article} 
\usepackage[accepted]{tmlr}

\usepackage{amsmath,amsfonts,bm}

\def\eqref#1{equation~\ref{#1}}

\def\1{\bm{1}}

\DeclareMathAlphabet{\mathsfit}{\encodingdefault}{\sfdefault}{m}{sl}
\SetMathAlphabet{\mathsfit}{bold}{\encodingdefault}{\sfdefault}{bx}{n}

\usepackage[dvipsnames,table,xcdraw]{xcolor}
\usepackage[pagebackref=true,breaklinks=true,colorlinks,citecolor=MidnightBlue]{hyperref}

\usepackage{hyperref}
\usepackage{url}

\usepackage{amsmath}
\usepackage{amssymb}
\usepackage{mathtools}
\usepackage{amsthm}

\usepackage{booktabs}
\usepackage{multirow}

\usepackage[utf8]{inputenc} 
\usepackage[T1]{fontenc}    
\usepackage{url}            
\usepackage{booktabs}       
\usepackage{amsfonts}       
\usepackage{nicefrac}       
\usepackage{microtype}      

\usepackage[frozencache,cachedir=minted-cache]{minted}
\usepackage[capitalize,noabbrev]{cleveref}

\theoremstyle{plain}
\newtheorem{theorem}{Theorem}[section]

\theoremstyle{definition}
\newtheorem{definition}[theorem]{Definition}

\theoremstyle{remark}

\usepackage{fvextra}
\DefineVerbatimEnvironment{jsoncode}{Verbatim}{
    breaklines=true,
    breaksymbolleft={},
    fontsize=\small
}

\usepackage[section]{placeins}
\usepackage{algpseudocode}
\usepackage[ruled,linesnumbered]{algorithm2e}
\SetKwComment{Comment}{/* }{ */}

\usepackage{bbding}

\usepackage[textsize=tiny]{todonotes}

\usepackage[misc]{ifsym}

\usepackage{enumitem}
\usepackage{soul}

\title{\raggedright Where Not to Learn: Prior-Aligned Training\\
with Subset-based Attribution Constraints}

\author{%
    \begin{minipage}{\textwidth}
    \raggedright
    {\name Ruoyu Chen$^1$, Shangquan Sun$^2$, Xiaoqing Guo$^3$, Kangwei Liu$^1$, Sanyi Zhang$^4$,\\
    Zhangcheng Wang$^5$, Shiming Liu$^6$, Qunli Zhang$^6$, Hua Zhang$^1$, Wei Wang$^7$, Xiaochun Cao$^{7,\textrm{\Letter}}$}\\[0.6em]
    {\addr $^1$University of Chinese Academy of Sciences, $^2$Nanyang Technological University,\\
    $^3$Hong Kong Baptist University, $^4$Communication University of China,\\
    $^5$ByteDance, $^6$Imperial College London, $^7$Sun Yat-sen University}\\[0.6em]
    {\small\normalfont
    \hypersetup{urlcolor=MidnightBlue}
    \textbf{First author:} \href{mailto:cryexplorer@gmail.com}{\nolinkurl{cryexplorer@gmail.com}}\\
    \textbf{\Letter\ Corresponding author:} \href{mailto:caoxiaochun@mail.sysu.edu.cn}{\nolinkurl{caoxiaochun@mail.sysu.edu.cn}}\\[0.4em]
    \raisebox{-0.1em}{\includegraphics[height=1em]{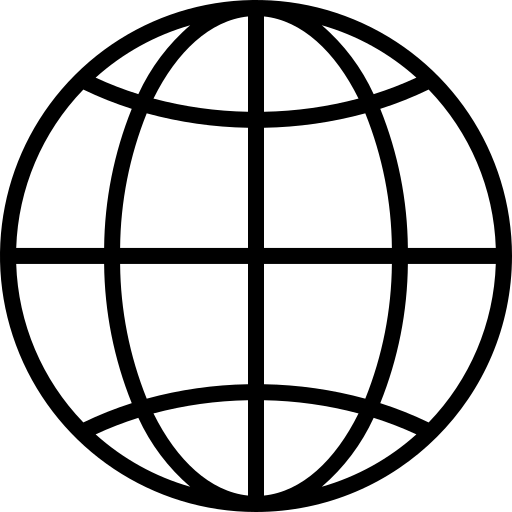}}\enspace
    \textbf{Project Page \& Code:} \url{https://Ruoyuchen10.github.io/Attribution-Constraint}}
    \end{minipage}%
}

\def\month{09}  
\def\year{2026} 
\def\openreview{{\hypersetup{urlcolor=MidnightBlue}\url{https://openreview.net/forum?id=2katjvBbIC}}} 

\begin{document}

\maketitle

\begin{abstract}
Reliable models should not only predict correctly, but also base their decisions on acceptable evidence. However, conventional supervised learning typically provides only class-level labels, allowing models to achieve high accuracy by exploiting shortcut correlations rather than intended decision evidence. Human priors, such as bounding boxes or target interface elements, can help constrain such behavior, but aligning model evidence with these priors remains challenging because learned decision evidence often diverges from human perception. In this work, we study attribution–prior alignment with subset-selection-based attribution. Motivated by prior deletion and insertion evaluations showing that subset-selection attribution can identify compact decision-supporting regions, we use it as a training-time signal to expose the model’s attributed evidence. When the top-attributed evidence deviates substantially from the prior region, we penalize off-prior attribution and encourage the model to shift its attributed evidence toward the intended regions. This yields a selective prior-constrained objective that avoids uniformly suppressing all non-prior regions. We validate our method on both image classification and click decision tasks in MLLM-based GUI agents. Across discriminative classification and autoregressive decision-making settings, our method improves task accuracy while enhancing attribution–prior alignment.
\end{abstract}

\section{Introduction}
\label{sec:intro}

\begin{figure}[!t]
    \centering  
    \includegraphics[width=0.7\textwidth]{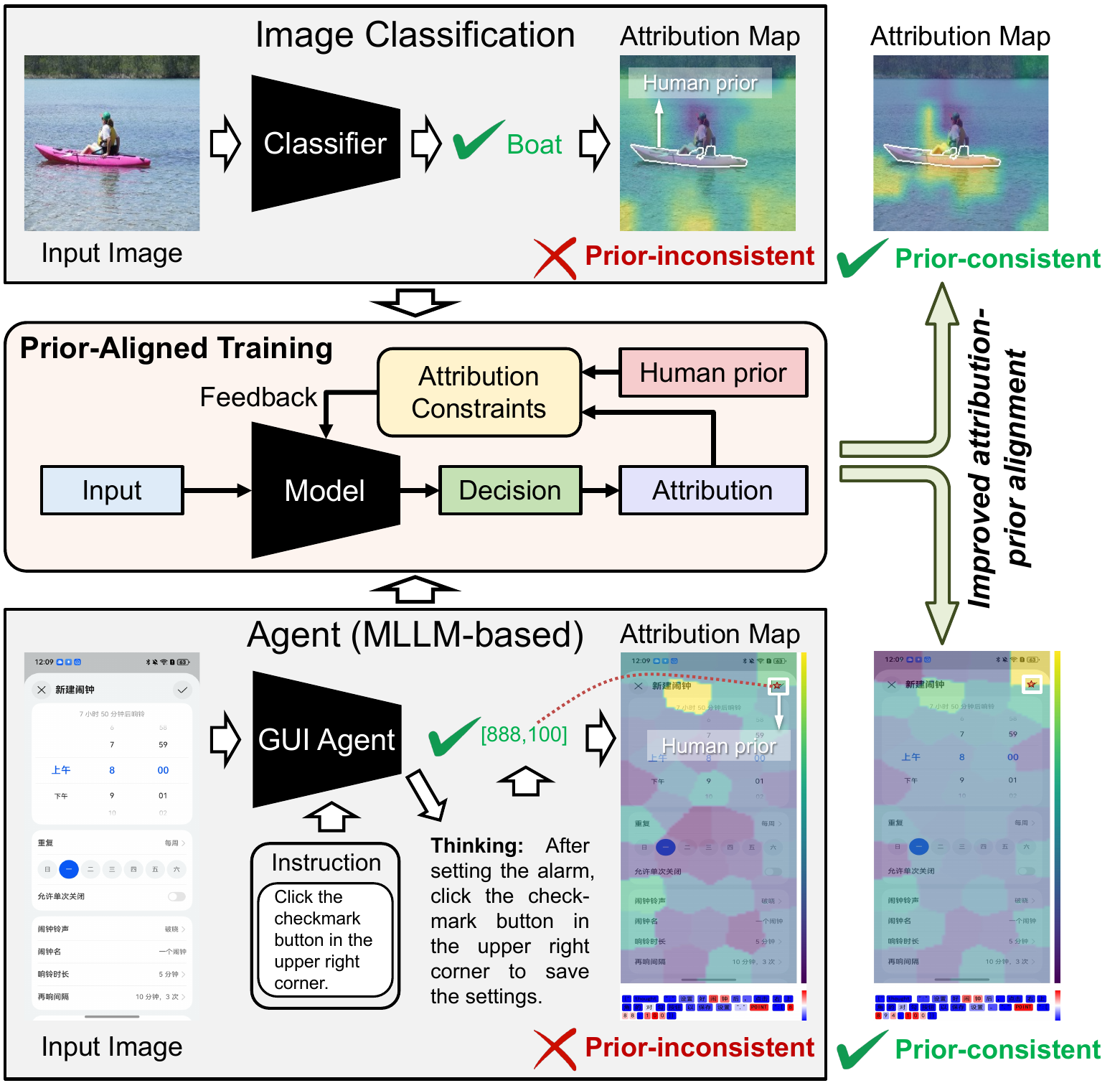} 
    \caption{Correct outputs do not guarantee prior-consistent attributed evidence: both a classifier and an MLLM-based GUI agent can succeed while their attribution maps deviate from specified human priors. We propose attribution-guided, prior-constrained alignment training to improve attribution--prior alignment and observed task performance.}
    \label{prior_align:motivation}
\end{figure}

Machine learning has recently achieved remarkable progress, with large-scale vision and multimodal models delivering strong performance across a wide range of tasks~\citep{li2025visual,li2025industryeqa,tu2026visualclaw}. As these models are increasingly deployed in real-world applications, understanding and constraining the evidence associated with their predictions becomes a central concern~\citep{kuznietsov2024explainable}. Models should not only predict correctly; their attributed evidence should also be consistent with acceptable and task-relevant cues. However, during training, models may learn shortcut correlations~\citep{geirhos2020shortcut,kauffmann2025explainable}, leading to seemingly correct outputs whose attributed evidence may be inconsistent with intended cues, warranting particular scrutiny in safety-critical or interactive settings, as shown in Fig.~\ref{prior_align:motivation}.

Standard supervised learning typically provides only class-level supervision, specifying what the correct output should be while leaving the input evidence associated with the prediction largely unconstrained~\citep{d2022underspecification,geirhos2018imagenet,rosenfeld2021risks}. As a result, even large-scale models may exploit the easiest or most statistically salient correlations rather than features specified by task semantics or annotator intent~\citep{d2022underspecification,turpin2023language}. Human priors can mitigate this issue by providing additional supervision about which input components are considered acceptable evidence. Here, human priors refer to human-recognizable cues about which input components (e.g., objects, regions, or attributes) should be emphasized for the prediction, typically provided as weak supervision such as sparse clicks, bounding boxes, or saliency annotations. We treat these priors as weak and potentially incomplete supervision rather than causal ground truth. However, aligning model attributions to such priors remains difficult, because learned attribution patterns often diverge from human perception~\citep{feather2019metamers,poursabzi2021manipulating,ngo2024alignment}.

Attribution methods~\citep{chen2024less,chen2025mllms} assign importance to input components associated with a model prediction. In this work, we adopt a perturbation-based operational criterion in which retaining or inserting important regions should better preserve the prediction, whereas removing them should reduce the corresponding prediction score. Prior attribution studies~\citep{chen2024less,chen2025interpreting,chen2025mllms} have shown that subset-selection-based attribution performs strongly under such deletion and insertion evaluations, suggesting that it can identify compact regions that support the model output under these perturbations. However, these attribution methods are primarily post-hoc explanations; how to use such attribution signals to align attributed evidence with specified human priors during training remains less explored. Some works~\citep{rao2023studying} leverage attribution signals for targeted model correction. RRR~\citep{ross2017right} and XIL~\citep{schramowski2020making} encourage explanations to agree with human-prior regions by suppressing gradients outside human-prior regions at the input or feature level, while MEGL~\citep{zhang2024megl} encourages feature activation maps to align with human annotations to learn more plausible evidence. However, these methods (i) operate on pointwise gradient- or activation-based explanations rather than explicitly searching for compact prediction-supporting subsets, and (ii) often apply hard, uniform suppression or enhancement, such as pushing all non-prior regions toward zero, without accounting for unequal contributions across regions.

Motivated by these challenges, we study how subset-selection attribution can be used as a training signal for attribution--prior alignment. Rather than proposing explanation alignment as a new paradigm, our goal is to instantiate attribution-guided alignment with subset-selection-based attribution and examine whether compact subset-based attributed evidence can support selective prior constraints across different prediction settings, as shown in Fig.~\ref{prior_align:motivation}. Accordingly, we frame the resulting evidence as attribution--prior alignment rather than as proof of causal grounding or deployment reliability.

We represent human priors as expected input regions, such as object bounding boxes or interface elements, and use subset-selection-based attribution~\citep{chen2024less,chen2025mllms} during training to identify compact regions that are sufficient to preserve the current prediction under the masking operation. We do not enforce a direct, global alignment to human priors. Instead, the deviation loss is activated only when the top-ranked region lies outside the prior, while a separate redundancy loss penalizes positive contributions from subsequently selected off-prior regions. Thus, agreement between the top-ranked region and the prior disables only the deviation term; later off-prior regions may still be constrained by the redundancy term according to their additional contributions. This yields a training objective that encourages prior-consistent attributed evidence while avoiding uniform suppression of all non-prior regions.

We evaluate the proposed framework across both image classification tasks and click decision tasks in multimodal large language model--based GUI (Graphical User Interface) agents~\citep{zhang2025agentcpm,han2026vlaa}. These experiments encompass both conventional discriminative prediction and autoregressive decision-making scenarios. To separate the effect of spatial supervision from that of the attribution objective, we include matched-supervision controls~\citep{stojanov2022learning,xiao2020noise} that receive the same masks or target boxes without using an attribution loss. For classification, we additionally evaluate all models with RISE~\citep{petsiuk2018rise} and gradient-based~\citep{selvaraju2020grad,zhao2024gradient} attribution families that are independent of the subset-selection attribution used during training. The classification results are reported over three independently seeded runs, whereas the limited 936-task GUI study is presented as a preliminary single-run proof of concept. Across the evaluated settings, our method improves observed task performance and aggregate attribution--prior alignment; improvements are also observed under the independent attribution evaluators in most classification model--evaluator combinations.

In summary, the contributions of this paper are:
\begin{itemize}
    \item We revisit attribution-guided human-prior alignment and instantiate it with subset-selection-based attribution, using compact prediction-supporting subsets as a training-time signal across discriminative and autoregressive prediction settings.
    
    \item We design a selective prior-constrained objective with independently applied deviation and redundancy terms: the deviation term is activated when the top-ranked region is off-prior, while the redundancy term constrains positive contributions from subsequently selected off-prior regions, thereby avoiding hard uniform suppression of all non-prior evidence.
    
    \item We evaluate the proposed instantiation on both image classification and MLLM-based GUI-agent click decision tasks, including explanation-guided baselines, supervision-matched controls, and attribution evaluators independent of the training-time attribution. The results demonstrate improved observed task performance and aggregate attribution--prior alignment, while the limited GUI study is characterized as a preliminary proof of concept.
\end{itemize}

\section{Related Work}

\textbf{Attribution methods.} Attribution methods aim to characterize the input evidence associated with a model prediction by assigning relevance to input components such as pixels, regions, or tokens. Existing approaches differ in mechanism, including gradient-based methods~\citep{selvaraju2020grad,zhao2024gradient,zhang2025redundancy,xing2025large}, perturbation-based methods~\citep{petsiuk2018rise,novello2022making}, Shapley value--based methods~\citep{lundberg2017unified,sun2023explain}, and attention-based methods~\citep{li2025token}. A central challenge is attribution faithfulness, namely, whether the attributed components capture prediction-sensitive evidence under a specified evaluation criterion. This has motivated sufficiency- and necessity-based formulations, in which retaining important regions should preserve the target prediction whereas removing them should reduce its score, as well as minimal-subset formulations that seek compact prediction-supporting evidence. Recent subset-selection-based methods~\citep{chen2024less,chen2025less,chen2025interpreting,chen2025mllms} have demonstrated strong performance under deletion and insertion evaluations by explicitly identifying compact prediction-supporting subsets. Accordingly, we use subset-selection attribution as the training-time signal for attribution--prior alignment.

\textbf{Attribution-guided learning} studies how attribution signals can be incorporated into training to shape model behavior beyond output supervision~\citep{gao2024going}. Some works encourage sparsity, smoothness, or consistency by regularizing gradient-based attributions during training, although predictive performance can be sensitive to the imposed regularization~\citep{erion2021improving,han2021explanation,pillai2022consistent}. Other works use counterfactual attribution for data augmentation to improve generalization~\citep{chen2025generalized,chen2025did}, but primarily target predictive generalization rather than explicit attribution--prior alignment. Explanation-guided methods incorporate human-provided annotations or explanation constraints to encourage model attributions to agree with intended input regions~\citep{ross2017right,schramowski2020making,selvaraju2019taking,zhang2023magi}. However, their training objectives generally operate on pointwise gradients, feature activation maps, or local surrogate explanations, such as Grad-CAM~\citep{selvaraju2020grad} and LIME~\citep{ribeiro2016should}, rather than explicitly searching for compact prediction-supporting subsets. They also often impose relatively uniform constraints over off-prior locations without accounting for the unequal contributions of different regions. In this paper, we instantiate attribution-guided prior alignment with the subset-selection-based LIMA~\citep{chen2024less,chen2025less} and EAGLE~\citep{chen2025mllms}. Their ordered attribution results allow us to selectively constrain off-prior regions according to their contributions to the current prediction, with evaluation focused on observed task performance and attribution--prior alignment.

\textbf{Spatial supervision.} Spatial annotations such as foreground masks and bounding boxes can be incorporated into training without explicitly constraining model attributions. \cite{xiao2020noise} disentangle foreground and background signals through foreground--background recombination, including Mixed-Rand, which places an object foreground on a randomly sampled background to weaken the original label--background correlation. \cite{stojanov2022learning} use foreground masks and multi-view correspondences to learn dense object descriptors, with random background removal encouraging more object-centric representations. In GUI-agent training, UI-R1~\citep{lu2025uir1} incorporates target-element bounding boxes into a rule-based coordinate reward for reinforcement learning, directly rewarding predicted clicks that fall within the target element. These approaches use spatial annotations through input augmentation or task-level rewards, but do not explicitly examine or constrain the attributed evidence associated with each prediction. In contrast, our method uses subset-selection attribution to determine when attributed evidence deviates from the specified prior and applies selective prior constraints accordingly. We further compare against supervision-matched augmentation and task-reward controls to distinguish the attribution objective from the benefit of receiving spatial annotations alone.

\section{Preliminaries and Problem Statement}

\subsection{Subset-selection based Attribution}\label{sec:attr}

Attribution methods seek to explain model decisions by quantifying the dependence of a prediction on individual input components. Subset-selection based attribution ranks sub-regions in the entire inputs by iteratively selecting compact decision-supporting subsets. Regions selected earlier are deemed more influential, defined as follows.

\begin{definition}[Subset-Selection-Based Attribution]
Given an input $\mathbf{x}$, a trained model $f$, and an objective set function $\mathcal{F}:2^{\mathcal{V}}\to\mathbb{R}$, subset-selection-based attribution sparsifies $\mathbf{x}$ into sub-regions $\mathcal{V}=\{v_1,\ldots,v_n\}$. For an ordering $\pi=(v_{\pi_1},\ldots,v_{\pi_n})\in\mathcal{P}(\mathcal{V})$, define its rank-$r$ prefix set as $S_r^\pi=\{v_{\pi_1},\ldots,v_{\pi_r}\}$. The target ordering is
\begin{equation}
    \pi^\star \in \underset{\pi \in \mathcal{P}(\mathcal{V})}{\arg\max}\; \sum_{r=1}^{|\mathcal{V}|} \mathcal{F}(S_r^\pi),
\end{equation}
where $\mathcal{P}(\mathcal{V})$ is the set of all permutations of $\mathcal{V}$. In practice, we approximately construct this ordering using greedy search or its accelerated variants.
\end{definition}

From this perspective, attribution is cast as a subset selection problem over $\mathcal{V}$, where decision evidence is characterized by compact, decision-supporting subsets and their induced ordering. 

\subsection{Problem Statement}

Attribution-based constrained training is formulated by introducing an attribution regularization term $\mathcal{L}_{\text{human}}$ that encourages consistency between model attributions and human priors. The resulting optimization objective is
\begin{equation}\small
    \min_{\theta}\mathbb{E}_{(x,y,H)\sim \mathcal{D}}\!\left[
    \underbrace{\mathcal{L}_{\text{task}}\!\left(f_{\theta}(x),y\right)}_{\text{task supervision}}
    + 
    \underbrace{\lambda\,\mathcal{L}_{\text{human}}\!\left(\operatorname{Attr}\!\left(f_{\theta}(x),y\right), H\right)}_{\text{human prior alignment}}
    \right],
\end{equation}
where $H$ denotes a human prior associated with sample $x$, and $\operatorname{Attr}$ is the attribution operator. We treat $H$ as weak, potentially incomplete guidance rather than causal ground truth. The objective therefore targets agreement with the specified prior and observed task performance; it does not imply that $H$ contains every legitimate feature or that attribution overlap establishes causal reliance.

\section{Method}

This section introduces our attribution-based prior-constrained alignment algorithm. Section~\ref{prior:principle} presents the alignment principle. Section~\ref{prior:loss} then details the loss-function instantiation. Finally, Section~\ref{prior:training} describes the overall training objective and optimization procedure.

\subsection{Evidence-Level Alignment Principle}\label{prior:principle}

We align model behavior with human priors by constraining attributed evidence rather than internal representations. This relies on subset-selection-based attribution, which identifies compact decision-supporting subsets and their induced ordering. Let $\mathcal{V}$ denote the set of input sub-regions and $H$ denote a human prior specified over the input space, such as bounding boxes or masks. For a given prediction, attribution ranks regions in $\mathcal{V}$ by their influence on the prediction. The top-ranked region and the higher-order regions are constrained by separate gates: the former activates the deviation term, whereas every off-prior region at rank $r\geq2$ is considered independently by the redundancy term.

Alignment is imposed asymmetrically: only off-prior attributed evidence is penalized, while prior-consistent regions remain unconstrained. We instantiate the same principle with LIMA~\cite{chen2024less,chen2025less} for image classifiers and EAGLE~\cite{chen2025mllms} for the autoregressive GUI agent. Both operate through model inputs and output probabilities, allowing a common loss construction without requiring the same internal architecture.

\subsection{Alignment with Subset-based Attribution}\label{prior:loss}

We instantiate the prior constrained training using the subset-based attribution framework in Section~\ref{sec:attr}, which produces an ordering over sub-regions $\mathcal{V}$ by decision influence.
Let $\pi = (v_{\pi_1}, v_{\pi_2}, \ldots, v_{\pi_{|\mathcal{V}|}})$ denote the ranking over sub-regions $\mathcal{V}$ induced by attribution, where regions appearing earlier are more influential, and let $S_r^\pi=\{v_{\pi_1},\ldots,v_{\pi_r}\}$ denote its rank-$r$ prefix set. Let $H$ denote the human prior specified over the input space (e.g., bounding boxes or masks). Since $H$ may not lie in the same discrete space as $\mathcal{V}$, we define an overlap function $\phi(v, H) \in [0,1]$, which measures the spatial consistency between a region $v$ and the human prior $H$ (e.g., IoU or mask coverage). A region is considered off-prior when $\phi(v, H)$ is small.

For a selected subset $S\subseteq\mathcal{V}$, let $m_S$ be the union of its binary region masks, $x_S=x\odot m_S$ the retained-subset input, and $x_{\mathcal{V}\setminus S}=x\odot(1-m_S)$ the complementary input. For classification, the target-class confidence is $p_\theta(y\mid x)=\operatorname{softmax}(f_\theta(x))_y$, and the LIMA set function is
\begin{equation}
\mathcal{F}^{\mathrm{cls}}_\theta(S)=s_{\mathrm{consistency}}(S)+s_{\mathrm{collaboration}}(\mathcal{V}\setminus S),
\end{equation}
where $s_{\mathrm{consistency}}(S)=p_\theta(y\mid x_S)$ and $s_{\mathrm{collaboration}}(\mathcal{V}\setminus S)=1-p_\theta(y\mid x_{\mathcal{V}\setminus S})$. Thus, retaining $S$ should preserve the target prediction, while removing it should reduce the corresponding confidence.

For the autoregressive GUI agent, let $\mathcal{I}$ denote the supervised token-position indices of the thinking--action sequence and $z_j$ the target token at index $j$. Following EAGLE, we use the analogous insight--necessity set function
\begin{equation}
\mathcal{F}^{\mathrm{gui}}_\theta(S)=s_{\mathrm{insight}}(S)+s_{\mathrm{necessity}}(\mathcal{V}\setminus S),
\end{equation}
with
\begin{align}
s_{\mathrm{insight}}(S)
&=\sum_{j\in\mathcal{I}}p_\theta(y_j=z_j\mid x_S,\mathrm{Prompt},y_{<j}),\\
s_{\mathrm{necessity}}(\mathcal{V}\setminus S)
&=\sum_{j\in\mathcal{I}}\left[1-p_\theta(y_j=z_j\mid x_{\mathcal{V}\setminus S},\mathrm{Prompt},y_{<j})\right].
\end{align}
The deviation and redundancy losses below use $\mathcal{F}^{\mathrm{cls}}_\theta$ for classifiers and $\mathcal{F}^{\mathrm{gui}}_\theta$ for the GUI agent.

\textbf{Deviation loss:} To prevent the most influential attribution region from deviating from the human prior during training, we introduce a \emph{Deviation Loss}. Since subset-selection–based attribution ranks regions according to a set function $\mathcal{F}(\cdot)$, deviations from the prior are addressed by suppressing the contribution of the top-ranked region. Specifically, when the most influential region $v_{\pi_1}$ exhibits low consistency with the human prior, we reduce its singleton-set utility $\mathcal{F}(\{v_{\pi_1}\})$ to discourage reliance on this region. The resulting optimization objective is
\begin{equation}
\mathcal{L}_{\text{deviation}}=\sum_{i=1}^{b}\mathcal{F}_\theta\!\left(\{v_{(i,\pi_1)}\}\right) \cdot \mathbf{1} \left[\phi(v_{(i,\pi_1)}, H_i) < \tau\right],
\end{equation}
where $b$ denotes the batch size, $v_{(i,\pi_1 )}$ is the most influential attribution region for the $i$-th sample, $\tau$ is a threshold determining consistency with the human prior, and $\mathbf{1}[\cdot]$ is the indicator function. No \emph{deviation} penalty is applied when the most influential region lies within the human prior; this gate does not disable the independently evaluated redundancy term. When the primary attribution region falls outside the prior, its utility score $\mathcal{F}$ is suppressed.

\textbf{Redundancy loss:} Beyond constraining the primary attribution region, we further regulate \emph{higher-order attribution regions}, referring to all attribution results beyond the top-ranked one. When such higher-order regions fall outside the human prior, their contribution to the model’s decision should be limited, as accumulating evidence from unintended regions leads to redundant and potentially spurious decision support. Intuitively, off-prior regions should not provide substantial additional gains once the primary evidence has been identified.
Since subset–based attribution constructs decision evidence sequentially via marginal gains of the set function $\mathcal{F}(\cdot)$, we suppress excessive marginal contributions from higher-order off-prior regions. This redundancy loss mitigates the accumulation effect in multi-region combinations, preventing off-prior regions from jointly contributing to the prediction and introducing shortcut cues. The objective is
\begin{equation}
    \mathcal{L}_{\text{redundancy}} = \sum_{i=1}^{b}\sum_{r=2}^{k} \mathsf{ReLU}\left( \Delta_{i,r} \right) \cdot \mathbf{1} \left[\phi(v_{(i,\pi_r)}, H_i) < \tau\right],
\end{equation}
For sample $i$, let $S_{i,r}^\pi=\{v_{(i,\pi_1)},\ldots,v_{(i,\pi_r)}\}$ be the rank-$r$ prefix set. Then $\Delta_{i,r} = \mathcal{F}_\theta(S_{i,r}^\pi) - \operatorname{sg}[\mathcal{F}_\theta(S_{i,r-1}^\pi)]$ denotes the marginal gain contributed by the region at rank $r$, $\operatorname{sg}[\cdot]$ is stop-gradient, and $k$ is the maximum number of selected regions. This term is evaluated for every rank $r\geq2$ independently of the top-region gate and penalizes only positive marginal gains from off-prior regions. Consequently, a prior-consistent top-ranked region incurs no deviation loss but does not exempt subsequent off-prior regions from the redundancy loss.





\begin{algorithm}[!htbp]
    \caption{Prior constrained training with subset-based attribution}
    \label{alg:alignment}
    \KwIn{Training data $(\mathbf{x}_i,y_i,H_i)$, alignment interval $T$, confidence threshold $\eta$, loss weights $\lambda_1,\lambda_2$, attribution length $k$.}
    \KwOut{Trained model parameters $\theta$.}

    Initialize model parameters $\theta$\;

    \For{$t=1$ \KwTo $T_{\max}$}{
        Sample a mini-batch $\{(\mathbf{x}_i, y_i, H_i)\}_{i=1}^{b}$\;
        Compute task loss $\mathcal{L}_{\mathrm{task}}$\;

        \uIf{$t \bmod T == 0$}{
            $\mathcal{L}_{\mathrm{deviation}} \gets 0$\;
            $\mathcal{L}_{\mathrm{redundancy}} \gets 0$\;

            \For{each sample $i$ in the batch}{
                \uIf{the prediction is correct and $p_\theta(y_i\mid\mathbf{x}_i)>\eta$}{
                    Compute top-$k$ attribution ranking $\pi$ using LIMA or EAGLE and detach the discrete search\;
                    If $\phi(v_{(i,\pi_1)},H_i)<\tau$, add $\mathcal{F}_\theta(\{v_{(i,\pi_1)}\})$ to $\mathcal{L}_{\mathrm{deviation}}$\;
                    For each $r=2,\ldots,k$, independently add $\mathsf{ReLU}(\Delta_{i,r})$ to $\mathcal{L}_{\mathrm{redundancy}}$ if $\phi(v_{(i,\pi_r)},H_i)<\tau$\;
                }
            }

            $\mathcal{L}_{\mathrm{total}} \gets
            \mathcal{L}_{\mathrm{task}}
            + \lambda_1 \mathcal{L}_{\mathrm{deviation}}
            + \lambda_2 \mathcal{L}_{\mathrm{redundancy}}$\;
        }
        \Else{
            $\mathcal{L}_{\mathrm{total}} \gets \mathcal{L}_{\mathrm{task}}$\;
        }

        Update model parameters $\theta$ using $\nabla \mathcal{L}_{\mathrm{total}}$\;
    }

    \Return $\theta$\;
\end{algorithm}

\subsection{Overall Training Objective}\label{prior:training}

We optimize the model using a mixed training objective that alternates between standard task supervision and evidence-level alignment. Specifically, alignment losses are applied only at regular intervals to reduce computational overhead and to avoid over-constraining the model during training.

Formally, let $t$ denote the training step and $T$ the alignment interval. When $t \bmod T = 0$, alignment is applied only to training samples that are correctly predicted by the model. For such samples, the optimization objective is
\begin{equation}
    \mathcal{L}
    = \mathcal{L}_{\text{task}}
    + \lambda_{1}\mathcal{L}_{\text{deviation}}
    + \lambda_{2}\mathcal{L}_{\text{redundancy}},
\end{equation}
where $\mathcal{L}_{\text{task}}$ denotes the standard task loss. For samples that are incorrectly predicted, as well as for all steps where $t \bmod T \neq 0$, the model is optimized using only the task loss.
The attribution search is a discrete outer selection step. The greedy region ranking, top-$k$ selection, overlap indicators, and confidence-based early stopping are computed without gradient tracking and are fixed for the subsequent update. Once the ranking and masks are fixed, gradients flow through the model scores $\mathcal{F}_\theta$ evaluated on the retained and complementary masked inputs, but not through the ranking, selection, threshold indicators, or stopping decision.
This intermittent and conditional strategy applies the alignment terms only to confidently correct training predictions, reducing computation and avoiding constraints on already incorrect outputs. The overall procedure is summarized in Algorithm~\ref{alg:alignment}.


\section{Experiments}

\subsection{Experimental Setup}

\textbf{Datasets.} We evaluate the proposed method on two representative tasks: image classification and a MLLM-based GUI agent clicking task.
For image classification, we use two datasets with high-quality object-level annotations. 
ImageNet-S~\citep{gao2022large} is a curated subset of ImageNet with 919 categories and pixel-level segmentation masks, while Saliency-Bench~\citep{zhang2025saliency} is constructed from MS COCO with object annotations. These datasets support evaluation of predictive performance and attribution overlap with a specified spatial prior.
For the GUI agent task, UI elements (e.g., buttons and icons) provide spatial priors over actionable targets. We collect 936 single-step Android clicking tasks with click locations and UI-element bounding boxes, enabling evaluation of task performance and attribution--prior overlap in a domain-specific MLLM. Because screenshots may contain sensitive visual content, any data release should document provenance and applicable collection authorization or consent, remove identifying information, and specify access conditions.

\textbf{Baselines.} We compare with representative attribution-based prior alignment baselines, including RRR~\citep{ross2017right}, which penalizes input-level gradients~\citep{simonyan2014deep} on non-prior regions, XIL~\citep{schramowski2020making}, which suppresses Grad-CAM~\citep{selvaraju2020grad} activations outside prior regions at the feature level, and MEGL~\citep{zhang2024megl}, which aligns Grad-CAM maps with mask annotations using an $\ell_1$ loss. For ViT-based architectures, Grad-CAM is replaced with Grad-ECLIP~\citep{zhao2024gradient}.

\textbf{Implementation Details.} For classification models, we compute attributions and perform prior alignment once every 10 training steps. For GUI-agent models, attributions are computed once every 5 steps. 
The loss balancing coefficients, $\lambda_1$ and $\lambda_2$, are both set to 0.5. During training, the subset-selection-based attribution sparsifies each image into 50 sub-regions. The attribution search selects at most 10 sub-regions and early-stops once the prediction confidence of the selected subset exceeds 0.8. We apply the attribution-alignment losses only to samples that are correctly predicted with confidence above 0.75, otherwise, the model is trained with the standard task loss only. More details please see the \textit{Appendix}.
Unless stated otherwise, classification results are reported as the mean and sample standard deviation over three independently seeded training runs. Each GUI method uses one training run; we therefore treat those results as descriptive, explicitly acknowledge the absence of training-run variance, and do not claim statistical significance.

\begin{table*}[!htbp]
    \caption{Image-classification performance and attribution--prior localization on Saliency-Bench and ImageNet-S. Point Game measures overlap with the specified mask, and Top-1 Acc. (PG$=1$) is conditioned on successful overlap. Results are mean $\pm$ sample standard deviation over three runs.}
    \begin{center}
        \resizebox{\textwidth}{!}{
            \begin{tabular}{ccccc|c|c|c}
                \toprule
                \textbf{Dataset} & \textbf{Human Prior} & \textbf{Model}  & \textbf{Methods} & \textbf{Attributions} & \textbf{Accuracy}& \textbf{Point Game} & \textbf{Top-1 Acc. (PG=1)} \\ 
                \midrule
                \multirow{15}{*}{Saliency-Bench} & \multirow{15}{*}{Masks} & \multirow{5}{*}{CLIP}
                & Fine-tuning & - & $0.6161\pm0.0044$ & $0.4567 \pm 0.0115$ & $0.8976 \pm 0.0150$ \\
                & & & RRR~\citep{ross2017right} & Input Gradient & $0.6250\pm0.0031$ & $0.4700 \pm 0.0200$ & $0.8725 \pm 0.0189$ \\
                & & & XIL~\citep{schramowski2020making} & Grad-ECLIP & $0.6173\pm0.0054$ & $0.4667 \pm 0.0153$ & $0.8862 \pm 0.0217$ \\
                & & & MEGL~\citep{zhang2024megl} & Grad-ECLIP & $0.6180\pm0.0053$ & $0.4700 \pm 0.0000$ & $0.8794 \pm 0.0325$ \\
                & & & \cellcolor[HTML]{D9D9D9}Ours & \cellcolor[HTML]{D9D9D9}LIMA & \cellcolor[HTML]{D9D9D9}$\mathbf{0.6543\pm0.0007}$ & \cellcolor[HTML]{D9D9D9}$\mathbf{0.4733\pm0.0153}$ & \cellcolor[HTML]{D9D9D9}$\mathbf{0.9154\pm0.0027}$ \\
                \cmidrule(l){3-8}
                
                & & \multirow{5}{*}{ViT (base)}
                & Fine-tuning & - & $0.5382\pm0.0122$ & $0.4933 \pm 0.0651$ & $0.8250 \pm 0.0109$ \\
                & & & RRR~\citep{ross2017right} & Input Gradient & $0.5359\pm0.0047$ & $0.4700 \pm 0.0624$ & $0.8226 \pm 0.0321$ \\
                & & & XIL~\citep{schramowski2020making} & Grad-ECLIP & $0.5204\pm0.0116$ & $0.4100 \pm 0.0520$ & $0.7378 \pm 0.0702$ \\
                & & & MEGL~\citep{zhang2024megl} & Grad-ECLIP & $0.5305\pm0.0291$ & $0.4167 \pm 0.0586$ & $0.7789 \pm 0.0749$ \\
                & & & \cellcolor[HTML]{D9D9D9}Ours & \cellcolor[HTML]{D9D9D9}LIMA & \cellcolor[HTML]{D9D9D9}$\mathbf{0.5806\pm0.0109}$ & \cellcolor[HTML]{D9D9D9}$\mathbf{0.5031\pm0.0100}$ & \cellcolor[HTML]{D9D9D9}$\mathbf{0.8276\pm0.0256}$ \\
                \cmidrule(l){3-8}
                
                & & \multirow{5}{*}{ResNet-101}
                & Fine-tuning & - & $0.4116\pm0.0094$ & $0.4533 \pm 0.0551$ & $0.6828 \pm 0.0224$ \\
                & & & RRR~\citep{ross2017right} & Input Gradient & $0.5567\pm0.0216$ & $0.5467 \pm 0.0208$ & $0.8725 \pm 0.0329$ \\
                & & & XIL~\citep{schramowski2020making} & Grad-CAM & $0.5536\pm0.0209$ & $0.5500 \pm 0.0100$ & $0.8606 \pm 0.0109$ \\
                & & & MEGL~\citep{zhang2024megl} & Grad-CAM & $0.5533\pm0.0190$ & $0.5571 \pm 0.0104$ & $\mathbf{0.8872 \pm 0.0434}$ \\
                & & & \cellcolor[HTML]{D9D9D9}Ours & \cellcolor[HTML]{D9D9D9}LIMA & \cellcolor[HTML]{D9D9D9}$\mathbf{0.5590\pm0.0180}$ & \cellcolor[HTML]{D9D9D9}$\mathbf{0.5600\pm0.0100}$ & \cellcolor[HTML]{D9D9D9}$0.8276\pm0.0256$\\
                
                \midrule

                \multirow{15}{*}{\begin{tabular}[c]{c}ImageNet-S\end{tabular}} & \multirow{15}{*}{Masks} & \multirow{5}{*}{CLIP} 
                & Fine-tuning & - & $0.7905\pm0.0004$ & $0.7633 \pm 0.0115$ & $\mathbf{0.8515 \pm 0.0067}$ \\
                & & & RRR~\citep{ross2017right} & Input Gradient & $0.7959\pm0.0012$ & $0.7633 \pm 0.0603$ & $0.8336 \pm 0.0217$ \\
                & & & XIL~\citep{schramowski2020making} & Grad-ECLIP & $0.7905\pm0.0007$ & $0.7633 \pm 0.0404$ & $0.8384 \pm 0.0010$ \\
                & & & MEGL~\citep{zhang2024megl} & Grad-ECLIP & $0.7925\pm0.0006$ & $0.7233 \pm 0.0153$ & $0.8433 \pm 0.0091$ \\
                
                & & & \cellcolor[HTML]{D9D9D9}Ours & \cellcolor[HTML]{D9D9D9}LIMA & \cellcolor[HTML]{D9D9D9}$\mathbf{0.7984\pm0.0003}$ & \cellcolor[HTML]{D9D9D9}$\mathbf{0.8233 \pm 0.0681}$ & \cellcolor[HTML]{D9D9D9}$0.8345 \pm 0.0142$ \\ 
                
                \cmidrule(l){3-8}
                
                & & \multirow{5}{*}{ViT (base)}
                
                & Fine-tuning & - & $0.7905\pm0.0031$ & $0.8700 \pm 0.0265$ & $0.8815 \pm 0.0212$ \\
                
                & & & RRR~\citep{ross2017right} & Input Gradient & $0.7917\pm0.0028$ & $0.9000 \pm 0.0100$ & $0.8630 \pm 0.0129$ \\
                
                & & & XIL~\citep{schramowski2020making} & Grad-ECLIP & $0.7007\pm0.0120$ & $0.9100 \pm 0.0100$ & $0.8242 \pm 0.0212$ \\
                
                & & & MEGL~\citep{zhang2024megl} & Grad-ECLIP & $0.7078\pm0.0072$ & $\mathbf{0.9133 \pm 0.0351}$ & $0.8355 \pm 0.0151$ \\
                
                & & & \cellcolor[HTML]{D9D9D9}Ours & \cellcolor[HTML]{D9D9D9}LIMA & \cellcolor[HTML]{D9D9D9}$\mathbf{0.7944\pm0.0033}$ & \cellcolor[HTML]{D9D9D9}${0.8767\pm0.0354}$ & \cellcolor[HTML]{D9D9D9}$\mathbf{0.8887\pm0.0190}$ \\
                
                \cmidrule(l){3-8}
                
                & & \multirow{5}{*}{ResNet-101}
                
                & Fine-tuning & - & $0.7030\pm0.0039$ & $0.8633 \pm 0.0153$ & $0.8383 \pm 0.0504$ \\
                
                & & & RRR~\citep{ross2017right} & Input Gradient & $0.7097\pm0.0024$ & $0.9000 \pm 0.0200$ & $0.7743 \pm 0.0374$ \\
                
                & & & XIL~\citep{schramowski2020making} & Grad-CAM & $0.7265\pm0.0013$ & $\mathbf{0.9200 \pm 0.0173}$ & $0.8259 \pm 0.0207$ \\
                
                & & & MEGL~\citep{zhang2024megl} & Grad-CAM & $0.7274\pm0.0025$ & $0.8967 \pm 0.0351$ & $0.7946 \pm 0.0385$ \\
                & & & \cellcolor[HTML]{D9D9D9}Ours & \cellcolor[HTML]{D9D9D9}LIMA & \cellcolor[HTML]{D9D9D9}$\mathbf{0.7314\pm0.0014}$ & \cellcolor[HTML]{D9D9D9}${0.8833 \pm 0.0503}$ & \cellcolor[HTML]{D9D9D9}$\mathbf{0.8449 \pm 0.0208}$ \\
                \bottomrule
            \end{tabular}
        }
    \end{center}
    \label{prior:image_classification_performance}
\end{table*}

\subsection{Evaluation on Image Classification}

We first validate our method on image classification tasks, where the selected datasets provide both class labels and object masks as human priors. 
In addition to comparing against direct fine-tuning to assess the benefit of prior supervision, we include attribution-based baselines that adopt different attribution methods and alignment strategies. 
We report top-1 accuracy and attribution--prior localization measured by Point Game~\citep{zhang2018top}, which tests whether the attributed region overlaps the annotated target object. Point Game measures localization consistency with the specified prior; by itself, it does not establish causal reliance or safe decision-making. We use LIMA~\citep{chen2024less,chen2025less} for the primary, common attribution evaluation and additionally test RISE and gradient-based evaluators below.

\textbf{Main Results.} As shown in Table~\ref{prior:image_classification_performance}, our method improves clean Top-1 accuracy over direct fine-tuning across CLIP~\citep{radford2021learning}, ViT~\citep{dosovitskiy2021image}, and ResNet~\citep{he2016deep} on both datasets. It also increases LIMA-based Point Game in the reported comparisons, indicating greater overlap between the attributed regions and the supplied masks. Accuracy conditioned on Point Game success describes performance on the subset whose attribution overlaps the prior; it should be interpreted as an association rather than evidence that the prior is the causal basis of the prediction.

\textbf{Independent attribution evaluation.} To reduce dependence on the attribution method used during training, we additionally evaluate Saliency-Bench models using the perturbation-based RISE~\citep{petsiuk2018rise} and gradient-based attribution---Grad-ECLIP~\citep{zhao2024gradient} for CLIP/ViT and Grad-CAM~\citep{selvaraju2020grad} for ResNet. Table~\ref{prior:independent_attribution} reports Point Game over three runs. Relative to fine-tuning, our method improves the independently evaluated score in five of the six model--evaluator combinations, with clear gains for ResNet (RISE: 40.33\% to 46.00\%; Grad-CAM: 37.90\% to 48.91\%). The CLIP Grad-ECLIP score is comparable but slightly lower (33.53\% versus 32.87\%). These results show that the observed attribution--prior improvement is not specific to LIMA, while attribution agreement alone still does not prove causal reliance.

\begin{table}[!htbp]
    \caption{Point Game (\%) on Saliency-Bench under attribution methods not used by our training objective, together with LIMA. Results are mean $\pm$ sample standard deviation over three runs. ``Gradient'' denotes Grad-ECLIP for CLIP/ViT and Grad-CAM for ResNet.}
    \begin{center}
        \resizebox{0.7\textwidth}{!}{
        \begin{tabular}{ccccc}
            \toprule
            \textbf{Model} & \textbf{Method} & \textbf{RISE} & \textbf{Gradient} & \textbf{LIMA} \\
            \midrule
            \multirow{5}{*}{CLIP}
            & Fine-tuning & $24.00\pm1.00$ & $\mathbf{33.53\pm0.59}$ & $45.67\pm1.15$ \\
            & RRR         & $\mathbf{25.33\pm1.53}$ & $32.52\pm0.81$ & $47.00\pm2.00$ \\
            & XIL         & $\mathbf{25.33\pm0.58}$ & $33.26\pm0.33$ & $46.67\pm1.53$ \\
            & MEGL        & $24.33\pm0.58$ & $32.80\pm0.58$ & $47.00\pm0.00$ \\
            & Ours        & $\mathbf{25.33\pm0.58}$ & $32.87\pm0.23$ & $\mathbf{47.33\pm1.53}$ \\
            \midrule
            \multirow{5}{*}{ViT (base)}
            & Fine-tuning & $38.00\pm2.65$ & $33.95\pm2.10$ & $49.33\pm6.51$ \\
            & RRR         & $37.00\pm2.00$ & $32.18\pm2.51$ & $47.00\pm6.24$ \\
            & XIL         & $\mathbf{44.00\pm3.00}$ & $29.82\pm11.06$ & $41.00\pm5.20$ \\
            & MEGL        & $43.33\pm4.73$ & $32.06\pm11.00$ & $41.67\pm5.86$ \\
            & Ours        & $42.00\pm2.65$ & $\mathbf{35.57\pm0.53}$ & $\mathbf{50.31\pm1.00}$ \\
            \midrule
            \multirow{5}{*}{ResNet-101}
            & Fine-tuning & $40.33\pm0.58$ & $37.90\pm1.22$ & $45.33\pm5.51$ \\
            & RRR         & $44.33\pm3.21$ & $48.30\pm0.27$ & $54.67\pm2.08$ \\
            & XIL         & $41.33\pm6.81$ & $47.96\pm0.41$ & $55.00\pm1.00$ \\
            & MEGL        & $42.00\pm3.61$ & $48.88\pm1.65$ & $55.71\pm1.04$ \\
            & Ours        & $\mathbf{46.00\pm2.65}$ & $\mathbf{48.91\pm0.47}$ & $\mathbf{56.00\pm1.00}$ \\
            \bottomrule
        \end{tabular}}
    \end{center}
    \label{prior:independent_attribution}
\end{table}

\textbf{Matched-supervision controls.} To distinguish subset-attribution constraints from generic access to spatial supervision, we add two controls that receive exactly the same foreground masks but use them only for input augmentation and optimize standard classification cross-entropy. Random Background Removal~\citep{stojanov2022learning} stochastically retains the annotated foreground and replaces the background with the preprocessing mean. Mixed-Rand~\citep{xiao2020noise} composites the target foreground with a foreground-free background sampled from another training image, reducing label--background correlation while retaining background diversity. Neither control uses an attribution loss or auxiliary localization objective. Table~\ref{prior:matched_supervision} reports clean Top-1 accuracy on the original validation images as mean $\pm$ sample standard deviation over three runs.

\begin{table}[!t]
    \caption{Clean Top-1 accuracy for matched-supervision controls on Saliency-Bench and ImageNet-S. Results are mean $\pm$ sample standard deviation over three runs.}
    \begin{center}
        \resizebox{\textwidth}{!}{
        \begin{tabular}{ccccc}
            \toprule
            \textbf{Dataset} & \textbf{Model} & \textbf{Random Background Removal} & \textbf{Mixed-Rand} & \textbf{Ours} \\
            \midrule
            \multirow{3}{*}{Saliency-Bench}
            & CLIP       & $0.6435\pm0.0031$ & $0.6478\pm0.0007$ & $\mathbf{0.6543\pm0.0007}$ \\
            & ViT (base) & $0.5745\pm0.0041$ & $\mathbf{0.5876\pm0.0088}$ & $0.5806\pm0.0109$ \\
            & ResNet-101 & $\mathbf{0.5629\pm0.0059}$ & $0.5591\pm0.0117$ & $0.5590\pm0.0180$ \\
            \midrule
            \multirow{3}{*}{ImageNet-S}
            & CLIP       & $0.8005\pm0.0005$ & $\mathbf{0.8062\pm0.0011}$ & $0.7984\pm0.0003$ \\
            & ViT (base) & $0.7694\pm0.0072$ & $0.7489\pm0.0021$ & $\mathbf{0.7944\pm0.0033}$ \\
            & ResNet-101 & $0.7076\pm0.0045$ & $0.6913\pm0.0010$ & $\mathbf{0.7314\pm0.0014}$ \\
            \bottomrule
        \end{tabular}}
    \end{center}
    \label{prior:matched_supervision}
\end{table}

Although no method dominates every setting, ours attains the highest average Top-1 accuracy across all six comparisons (0.6864, versus 0.6764 for Random Background Removal and 0.6735 for Mixed-Rand). The advantage is largest in aggregate on ImageNet-S, where the averages are 0.7747 versus 0.7592 and 0.7488; on ViT and ResNet-101, our method exceeds the strongest matched control by 2.50 and 2.38 percentage points, respectively. Thus, mask access and mask-guided augmentation alone do not fully account for the aggregate gains.

\begin{figure*}[!htbp]
    \centering  
    \includegraphics[width=\textwidth]{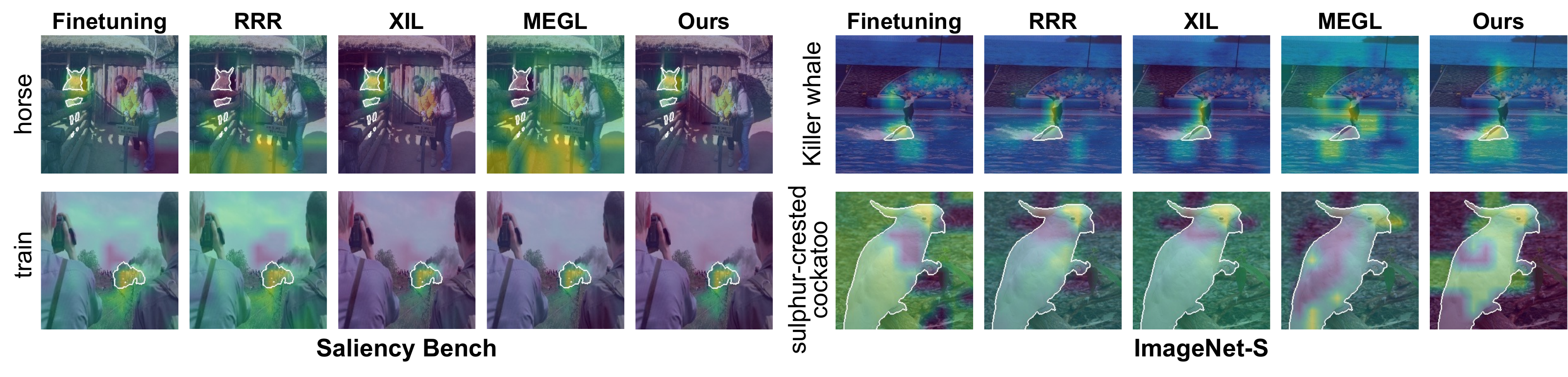} 
    \caption{Qualitative comparison on Saliency-Bench and ImageNet-S. For each method, we visualize LIMA-based attributions on the same inputs; white masks indicate human priors (target object regions).}
    \label{prior_align:visualization_classification}
\end{figure*}

Figure~\ref{prior_align:visualization_classification} shows qualitative LIMA attributions for all models. In images with multiple co-occurring objects, our method produces maps that are more concentrated on the supplied target masks than the displayed baselines. This visual localization is consistent with the Point Game results but does not by itself demonstrate reduced causal reliance on background content.

\subsection{Ablation Study}

\textbf{Ablation of the components.} Table~\ref{prior:ablation_components} presents ablations of the deviation loss and redundancy loss across different backbones on the Saliency-Bench dataset. Overall, the deviation loss brings consistent gains in both accuracy and decision reasonability. This indicates that explicitly penalizing reliance on off-prior evidence can effectively steer the model to ground its most influential evidence on human-recognized regions, which improves not only prediction performance but also prior-consistent explanations. In contrast, the redundancy loss mainly affects the quality of the explanation: when combined with the deviation loss, it yields an additional (typically mild) improvement in Point Game, while its impact on accuracy is limited. This behavior aligns with its design goal, by accounting for the cumulative effect of selected regions, the redundancy term suppresses repeated/overlapping evidence and encourages more efficient evidence allocation, thereby slightly enhancing decision reasonability.

\begin{table}[!htbp]
    \caption{Ablation studies on the deviation loss and the redundancy loss on the Saliency Bench dataset.}
    \begin{center}
        \resizebox{0.7\textwidth}{!}{
            \begin{tabular}{c|cc|cc}
            \toprule
            \textbf{Models}                  & \textbf{Deviation loss} & \textbf{Redundancy loss} & \textbf{Accuracy} & \textbf{Point Game} \\ \midrule
            \multirow{3}{*}{CLIP}   &  \XSolidBrush & \XSolidBrush  & 0.6076   & 0.4567     \\
                                    &  \Checkmark  & \XSolidBrush & 0.6525   & 0.4660     \\
                                    & \Checkmark  & \Checkmark & \textbf{0.6551}   & \textbf{0.4733}     \\ \midrule
            \multirow{3}{*}{ViT}    & \XSolidBrush & \XSolidBrush & 0.5150    & 0.4933     \\
                                    & \Checkmark  & \XSolidBrush & 0.5359   & 0.4806     \\
                                    & \Checkmark  & \Checkmark & \textbf{0.5690}    & \textbf{0.5031}     \\ \midrule
            \multirow{3}{*}{ResNet} & \XSolidBrush & \XSolidBrush & 0.4062   & 0.4533     \\
                                    & \Checkmark  & \XSolidBrush & 0.5535   & 0.5465     \\
                                    &  \Checkmark  & \Checkmark  & \textbf{0.5590}    & \textbf{0.5600}     \\ \bottomrule
            \end{tabular}
        }
    \end{center}
    \label{prior:ablation_components}
\end{table}

\begin{figure}[!htbp]
    \centering  
    \includegraphics[width=0.6\textwidth]{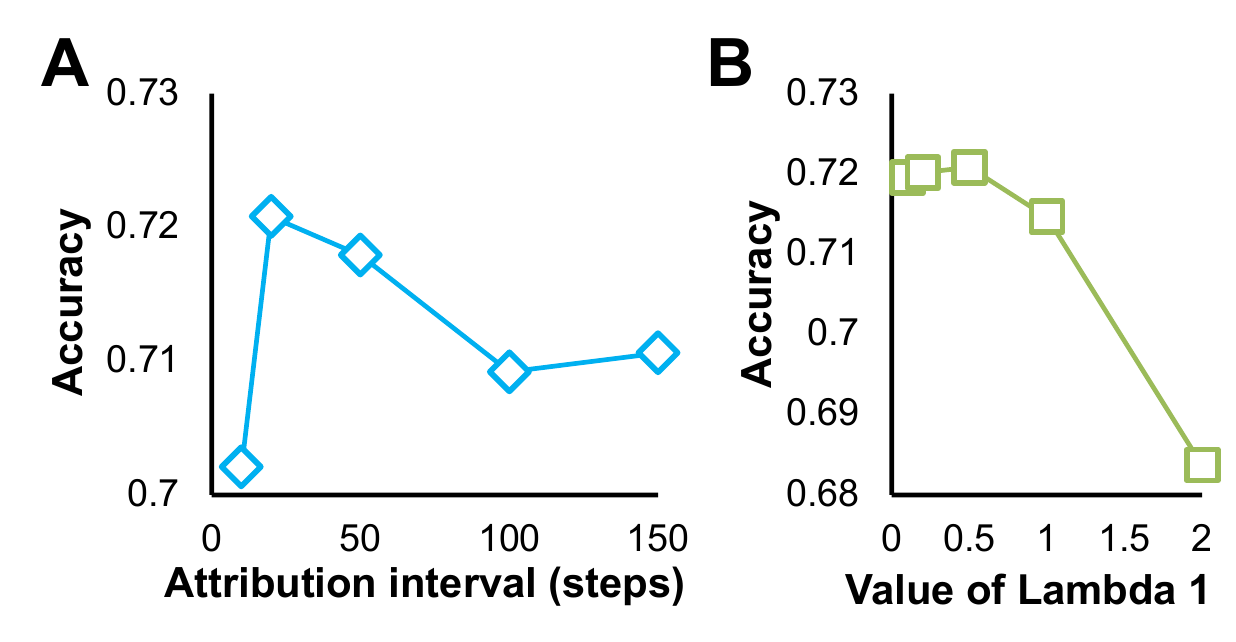} 
    \caption{Impact of training hyperparameters on model performance. \textbf{A.} Effect of the attribution interval on validation accuracy. \textbf{B.} Effect of the loss balancing coefficient $\lambda_1$ on validation accuracy.}
    \label{prior_align:sensitivity}
\end{figure}

\textbf{Parameter sensitivity analysis.} We conduct a parameter sensitivity study on ImageNet-S using the ViT backbone, focusing on (i) the step interval for applying attribution-based prior constraints during training and (ii) the weighting coefficient of the deviation loss. Figure~\ref{prior_align:sensitivity}A shows that applying the constraint more frequently (i.e., using a smaller interval) can improve classification accuracy, but enforcing it too often may disrupt optimization of the primary task and degrade performance. Figure~\ref{prior_align:sensitivity}B indicates that accuracy remains stable for small-to-moderate $\lambda_1$, while overly large $\lambda_1$ causes a clear drop, suggesting that the deviation loss should be weighted moderately to avoid overwhelming the main objective.

\textbf{Performance under Gaussian noise.} Table~\ref{prior:noise} reports a single paired run (seed 0) measuring both absolute corrupted accuracy and degradation from clean accuracy on Saliency-Bench under Gaussian noise with $\sigma=0.1$. Our method has higher noisy accuracy than fine-tuning for CLIP (61.57\% versus 59.95\%) and ResNet (45.72\% versus 12.15\%). However, relative degradation is smaller only for ResNet; for CLIP, our method starts from a higher clean accuracy but undergoes a larger drop. We therefore interpret these results as a descriptive diagnostic of improved absolute performance under this corruption, with mixed relative stability, rather than uniformly stronger noise robustness.

\begin{table}[!htbp]
    \caption{Single-run (seed 0) clean and Gaussian-noise ($\sigma=0.1$) Top-1 accuracy on Saliency-Bench, together with absolute and relative degradation from clean performance.}
    \begin{center}
        \resizebox{0.7\textwidth}{!}{
            \begin{tabular}{cc|cccc}
            \toprule
            \textbf{Model} & \textbf{Method} & \textbf{Clean Acc.} & \textbf{Noisy Acc.} & \textbf{Absolute Drop} & \textbf{Relative Drop} \\ \midrule
            \multirow{2}{*}{CLIP}   & Fine-tuning & 0.6076 & 0.5995 & 0.0081 & 1.33\% \\
                                    & Ours        & \textbf{0.6551} & \textbf{0.6157} & 0.0394 & 6.01\% \\ \midrule
            \multirow{2}{*}{ResNet} & Fine-tuning & 0.4062 & 0.1215 & 0.2847 & 70.09\% \\
                                    & Ours        & \textbf{0.5590} & \textbf{0.4572} & 0.1018 & 18.21\% \\ \bottomrule
            \end{tabular}
        }
    \end{center}
    \label{prior:noise}
\end{table}

\FloatBarrier
\subsection{Extension to MLLM-based GUI Agent}

Next, we validate our method in a GUI agent setting. We use AgentCPM-GUI~\citep{zhang2025agentcpm}, a reasoning-oriented MLLM that produces both \texttt{thinking} and a final \texttt{decision} (click action). We adopt supervised fine-tuning (SFT) as the primary training paradigm, using data that contains the target decision together with GPT-distilled \texttt{thinking} traces. Our goal is to improve the consistency between the decision evidence expressed in \texttt{thinking} and the executed \texttt{decision} via attribution-based consistency regularization. We employ EAGLE~\citep{chen2025mllms} for attribution, which is tailored to MLLMs. The classification baselines do not transfer directly to this setting: their standard formulations act on a single target-class score, input gradient, or designated feature map, whereas the GUI agent generates a multi-token thinking--action sequence. We therefore add two non-attribution controls in addition to standard SFT. Inspired by foreground--background intervention and background-randomization controls~\citep{xiao2020noise,stojanov2022learning}, \emph{Box-Guided Augmentation SFT} receives the same target bounding boxes as our method but uses them only to randomly mask or blur off-box pixels while optimizing ordinary SFT cross-entropy. \emph{Task-Reward GRPO} follows GRPO~\citep{shao2024deepseekmath} and its rule-reward adaptations to GUI action prediction~\citep{lu2025uir1,zhang2025agentcpm}, but uses only a task reward for valid structured output, the correct action type, and click success or distance to the target box; it does not use EAGLE, Point Game, or any attribution reward. The evaluation metrics and control details are described in Appendix~\ref{prior:gui_evaluation_metrics}.

\begin{figure*}[!htb]
    \centering  
    \includegraphics[width=\textwidth]{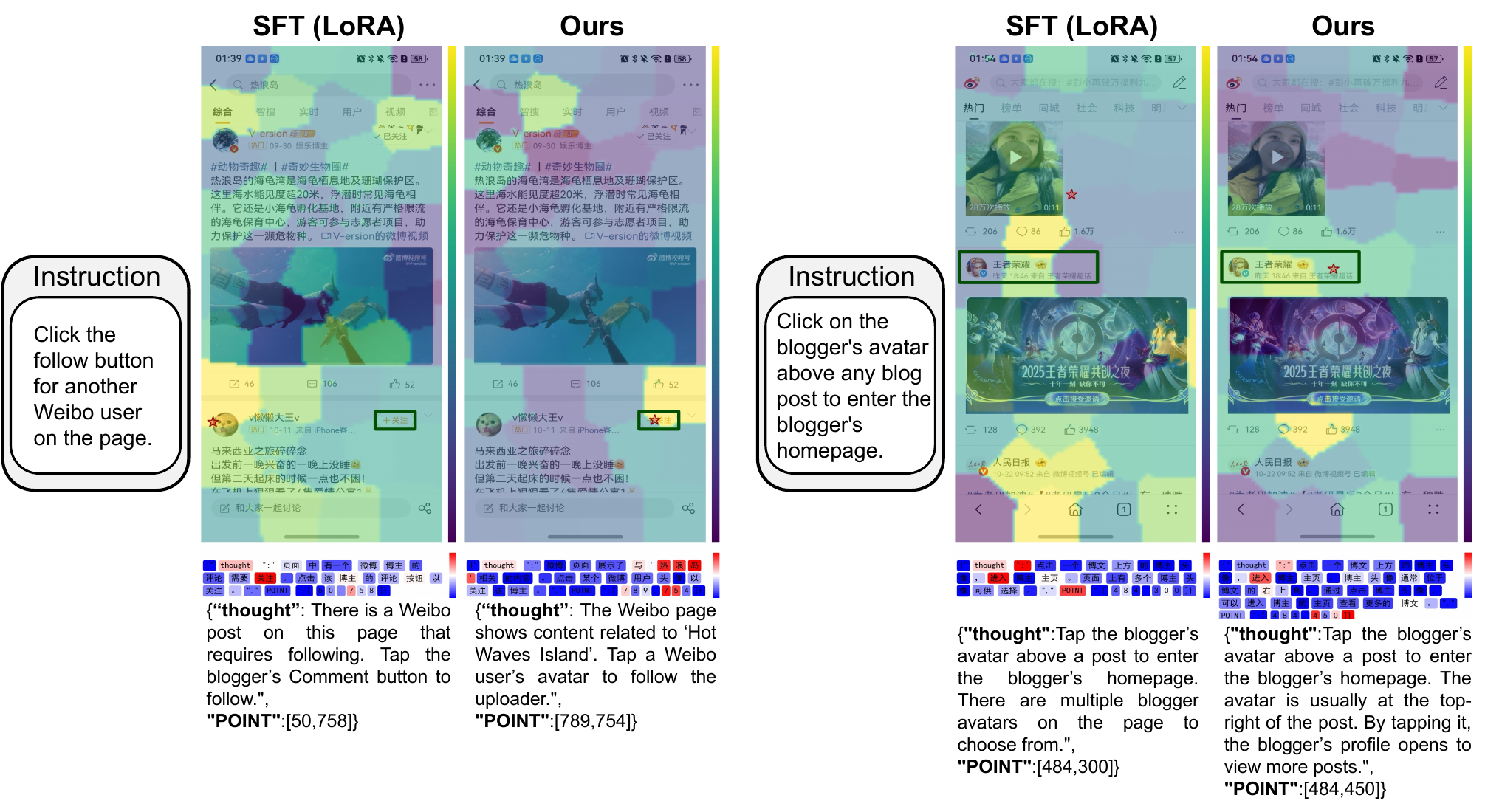} 
    \caption{Qualitative GUI examples comparing SFT and our method, with attribution heatmaps, predicted click locations (stars), and target UI bounding boxes used as spatial priors.}
    \label{prior_align:gui_task_success}
\end{figure*}

Table~\ref{prior:gui-agent_performance} summarizes the single-run GUI results. Standard SFT obtains 84.61\% click success and a distance error of 94.71. Box-Guided Augmentation SFT improves these values to 86.15\% and 85.26, while task-reward GRPO reaches 87.69\% and 78.42. Our attribution-constrained SFT obtains the highest click success, 89.23\%, and a distance error of 78.64; GRPO has a marginally lower distance error by 0.22. Thus, both additional spatial supervision and task-reward optimization provide useful controls, but neither matches our click-success gain. These values come from one training run per method and are descriptive rather than evidence of statistical significance or deployment reliability. Figure~\ref{prior_align:gui_task_success} shows representative examples.

\begin{table*}[!htb]
    \caption{Single-run evaluation on the GUI clicking task with AgentCPM-GUI. Box-Guided Augmentation SFT and task-reward GRPO use the target boxes during training but no attribution objective. Click success and distance error measure observed task performance; Point Game measures localization with the target UI prior, and the final two metrics condition on PG$=1$.}
    \begin{center}
        \resizebox{\textwidth}{!}{
            \begin{tabular}{c|cc|c|cc}
            \toprule
            \multirow{2}{*}{\textbf{Methods}} & \multicolumn{2}{c|}{\textbf{Task Performance}} & \multirow{2}{*}{\textbf{Point Game ($\uparrow$)}} & \multicolumn{2}{c}{\textbf{PG-conditioned Metrics}}            \\ 
             & Click success rate ($\uparrow$) & Distance error ($\downarrow$) & & Click success rate (PG=1) ($\uparrow$) & Distance error (PG=1) \\ \midrule
            SFT (LoRA) & 84.61\% & 94.71 & 0.8153 & 96.22\% & 7.11 \\
            Box-Guided Augmentation SFT (LoRA) & 86.15\% & 85.26 & 0.8462 & 96.36\% & 0.57 \\
            Task-Reward GRPO (LoRA) & 87.69\% & \textbf{78.42} & 0.7692 & 98.00\% & 1.42 \\
            \cellcolor[HTML]{D9D9D9}Ours (LoRA) & \cellcolor[HTML]{D9D9D9}\textbf{89.23\%} & \cellcolor[HTML]{D9D9D9}78.64 & \cellcolor[HTML]{D9D9D9}\textbf{0.8615} & \cellcolor[HTML]{D9D9D9}\textbf{100\%} & \cellcolor[HTML]{D9D9D9}\textbf{0.0} \\ \bottomrule
            \end{tabular}
        }
    \end{center}
    \label{prior:gui-agent_performance}
\end{table*}

\begin{figure}[!htb]
    \centering  
    \includegraphics[width=\textwidth]{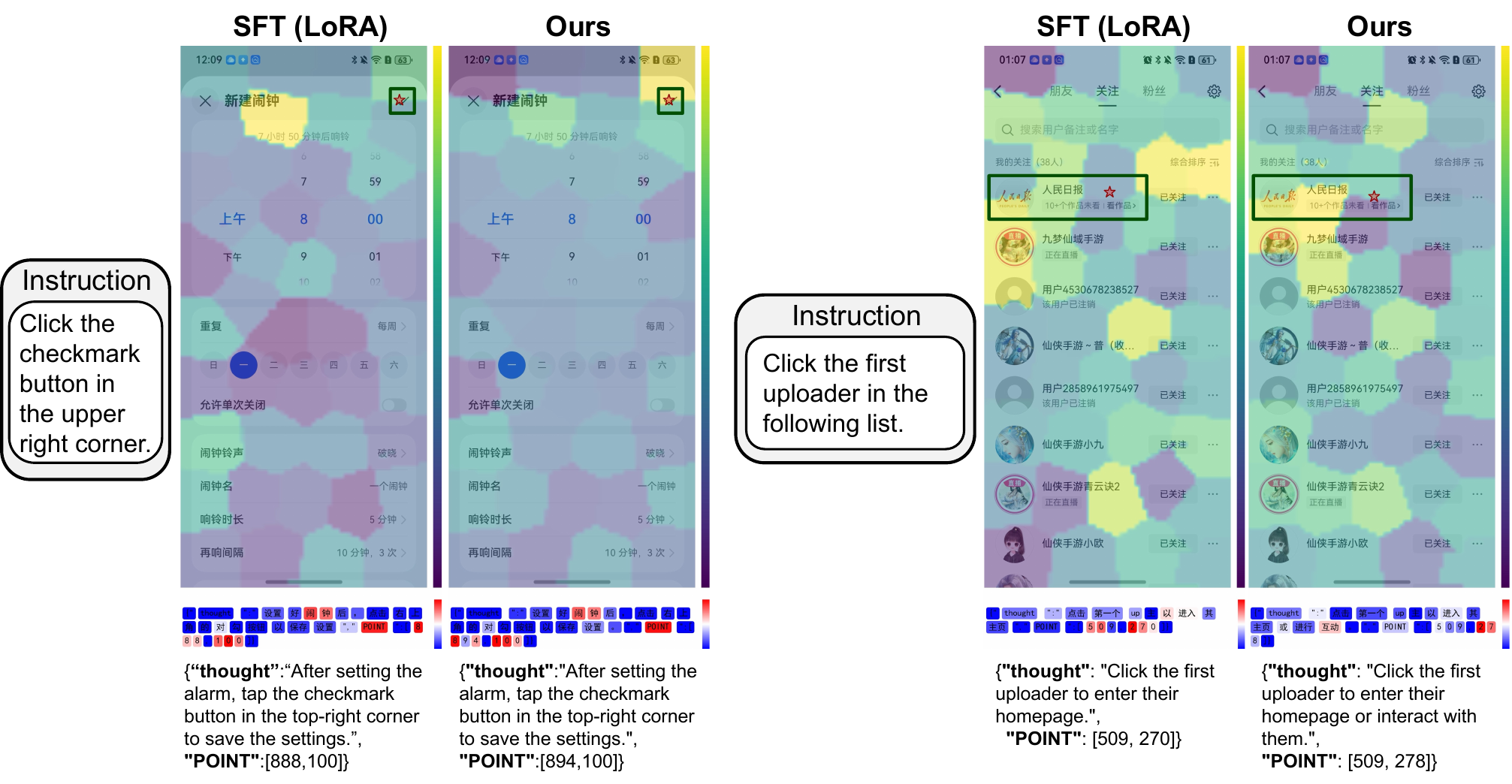} 
    \caption{GUI clicking example comparing SFT and our method, showing attribution maps and correct clicks on the target checkmark.}
    \label{prior_align:gui_attribution_compare}
\end{figure}

\begin{samepage}
We next evaluate EAGLE attribution for the generated thinking--action sequence. Point Game is 0.8153 for SFT, 0.8462 for Box-Guided Augmentation SFT, 0.7692 for task-reward GRPO, and 0.8615 for our method. The box-guided control shows that direct spatial augmentation can improve attribution--prior overlap to some extent. In contrast, GRPO improves click performance but lowers Point Game, showing that optimizing a verifiable click reward does not necessarily align the attributed thinking--action evidence with the target UI region. Our method obtains the highest Point Game while also obtaining the highest click success. Figure~\ref{prior_align:gui_attribution_compare} provides a qualitative comparison in which SFT and our method both click correctly but emphasize different screen regions.

\end{samepage}

The PG$=1$ columns are computed on method-dependent subsets and therefore should not be compared as population-level reliability estimates. In particular, the conditioned click-success and distance-error values describe an association between attribution--prior overlap and task performance; they do not show that prior overlap causes successful clicks or guarantees reliability.

\begin{figure}[!htb]
    \centering  
    \includegraphics[width=0.48\textwidth]{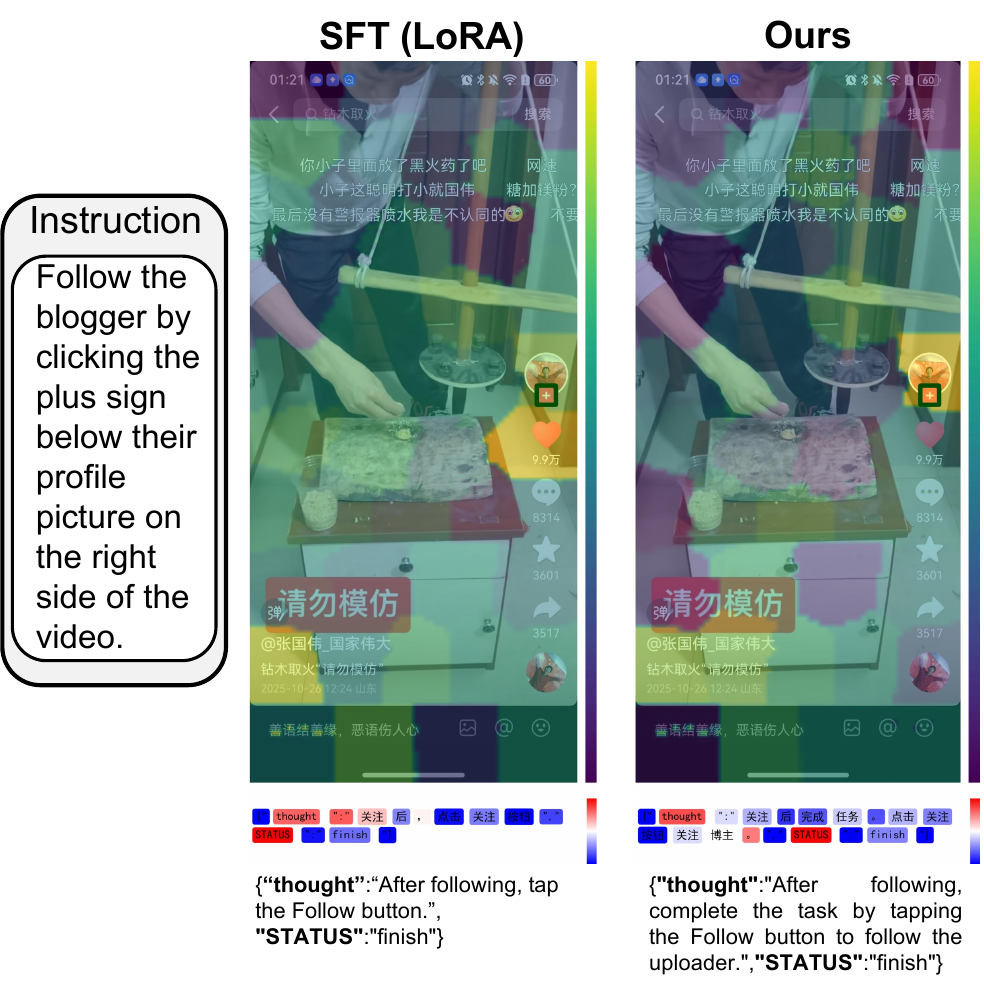} 
    \caption{Failure-case comparison on the GUI agent task.}
    \label{prior_align:gui_failure}
\end{figure}

\textbf{Failure analysis.} In Figure~\ref{prior_align:gui_failure}, neither model outputs a \texttt{POINT}; both return a \texttt{STATUS}. Their attributions nevertheless differ: SFT emphasizes the \emph{like} button, whereas our method emphasizes the intended \emph{follow} button. This qualitative example shows improved localization to the specified target despite an incorrect final action format, but it is not evidence of causal grounding.

\section{Conclusion}

We presented a prior-aligned training framework that periodically constrains subset-based attributions using human-provided object masks or UI-element bounding boxes. The deviation and redundancy terms act on different ranks of the attribution ordering, allowing the method to penalize influential off-prior regions without uniformly suppressing all context. Across image classification and GUI clicking tasks, the method improves observed task performance and increases attribution overlap with the specified prior; the classification improvement is also visible under several attribution methods not used during training. These findings support attribution--prior alignment, but Point Game and cross-attribution agreement do not establish causal grounding, distribution-shift robustness, or deployment safety.

\subsubsection*{Broader Impact Statement}

This work relates to \emph{explanation-guided learning}, which incorporates spatial priors in addition to output labels. Such supervision may help auditing and debugging by making a specified notion of acceptable evidence explicit. However, attribution--prior overlap must not be interpreted as proof that a model is causally grounded, robust, or safe, especially in high-stakes domains or deployed GUI agents.

\textbf{Risks and safeguards.} Human masks and UI-region annotations can encode annotator bias, omit legitimate contextual evidence, and disadvantage users or interfaces whose languages, visual conventions, or accessibility settings differ from the training data. Erroneous prior constraints may suppress valid evidence, while high overlap scores may create unwarranted confidence. GUI screenshots can also contain sensitive information, and autonomous clicks may trigger consequential actions. Data releases should therefore document provenance and applicable consent or authorization, minimize and de-identify sensitive content, and use access controls where needed. Deployment safeguards should include uncertainty reporting, restricted action permissions, human confirmation for consequential operations, and evaluation across diverse interfaces and accessibility settings.



\bibliography{main}

@String(CVPR  = {\textit{CVPR}})

@String(ICCV  = {\textit{ICCV}})

@String(NIPS  = {\textit{NeurIPS}})

@String(BMVC  =	{\textit{BMVC}})

@String(ICML  = {\textit{ICML}})

@String(ICLR  = {\textit{ICLR}})

@String(IJCAI = {\textit{IJCAI}})

@String(NAACL = {\textit{NAACL}})

@String(AAAI = {\textit{AAAI}})

@String(SIGKDD = {\textit{SIGKDD}})

@String(JMLR = {Journal of Machine Learning Research})

@String(TPAMI  = {IEEE Transactions on Pattern Analysis and Machine Intelligence})

@String(IJCV  = {International Journal of Computer Vision})

@article{li2025visual,
  title={Visual large language models for generalized and specialized applications},
  author={Li, Yifan and Lai, Zhixin and Bao, Wentao and Tan, Zhen and Dao, Anh and Sui, Kewei and Shen, Jiayi and Liu, Dong and Liu, Huan and Kong, Yu},
  journal={arXiv preprint arXiv:2501.02765},
  year={2025}
}

@inproceedings{li2025industryeqa,
  title={IndustryEQA: Pushing the Frontiers of Embodied Question Answering in Industrial Scenarios},
  author={Li, Yifan and Chen, Yuhang and Dao, Anh and Li, Lichi and Cai, Zhongyi and Tan, Zhen and Chen, Tianlong and Kong, Yu},
  booktitle=NIPS,
  year={2025}
}

@article{kuznietsov2024explainable,
  title={Explainable AI for safe and trustworthy autonomous driving: A systematic review},
  author={Kuznietsov, Anton and Gyevnar, Balint and Wang, Cheng and Peters, Steven and Albrecht, Stefano V},
  journal={IEEE Transactions on Intelligent Transportation Systems},
  volume={25},
  number={12},
  pages={19342-19364},
  year={2024}
}

@article{geirhos2020shortcut,
  title={Shortcut learning in deep neural networks},
  author={Geirhos, Robert and Jacobsen, J{\"o}rn-Henrik and Michaelis, Claudio and Zemel, Richard and Brendel, Wieland and Bethge, Matthias and Wichmann, Felix A},
  journal={Nature Machine Intelligence},
  volume={2},
  number={11},
  pages={665--673},
  year={2020}
}

@article{kauffmann2025explainable,
  title={Explainable AI reveals Clever Hans effects in unsupervised learning models},
  author={Kauffmann, Jacob and Dippel, Jonas and Ruff, Lukas and Samek, Wojciech and M{\"u}ller, Klaus-Robert and Montavon, Gr{\'e}goire},
  journal={Nature Machine Intelligence},
  pages={1--11},
  year={2025}
}

@article{d2022underspecification,
  title={Underspecification presents challenges for credibility in modern machine learning},
  author={D'Amour, Alexander and Heller, Katherine and Moldovan, Dan and Adlam, Ben and Alipanahi, Babak and Beutel, Alex and Chen, Christina and Deaton, Jonathan and Eisenstein, Jacob and Hoffman, Matthew D and others},
  journal=JMLR,
  volume={23},
  number={226},
  pages={1--61},
  year={2022}
}

@inproceedings{geirhos2018imagenet,
  title={ImageNet-trained CNNs are biased towards texture; increasing shape bias improves accuracy and robustness},
  author={Geirhos, Robert and Rubisch, Patricia and Michaelis, Claudio and Bethge, Matthias and Wichmann, Felix A and Brendel, Wieland},
  booktitle=ICLR,
  year={2018}
}

@inproceedings{rosenfeld2021risks,
  title={The Risks of Invariant Risk Minimization},
  author={Rosenfeld, Elan and Ravikumar, Pradeep Kumar and Risteski, Andrej},
  booktitle=ICLR,
  year={2021}
}

@inproceedings{turpin2023language,
  title={Language models don't always say what they think: Unfaithful explanations in chain-of-thought prompting},
  author={Turpin, Miles and Michael, Julian and Perez, Ethan and Bowman, Samuel},
  booktitle=NIPS,
  pages={74952--74965},
  year={2023}
}

@inproceedings{feather2019metamers,
  title={Metamers of neural networks reveal divergence from human perceptual systems},
  author={Feather, Jenelle and Durango, Alex and Gonzalez, Ray and McDermott, Josh},
  booktitle=NIPS,
  pages={10078--10089},
  year={2019}
}

@inproceedings{poursabzi2021manipulating,
  title={Manipulating and measuring model interpretability},
  author={Poursabzi-Sangdeh, Forough and Goldstein, Daniel G and Hofman, Jake M and Wortman Vaughan, Jennifer Wortman and Wallach, Hanna},
  booktitle={Proceedings of the 2021 CHI conference on human factors in computing systems},
  pages={1--52},
  year={2021}
}

@inproceedings{ngo2024alignment,
  title={The Alignment Problem from a Deep Learning Perspective},
  author={Ngo, Richard and Chan, Lawrence and Mindermann, S{\"o}ren},
  booktitle=ICLR,
  year={2024}
}

@inproceedings{radford2021learning,
  title={Learning transferable visual models from natural language supervision},
  author={Radford, Alec and Kim, Jong Wook and Hallacy, Chris and Ramesh, Aditya and Goh, Gabriel and Agarwal, Sandhini and Sastry, Girish and Askell, Amanda and Mishkin, Pamela and Clark, Jack and others},
  booktitle=ICML,
  pages={8748--8763},
  year={2021}
}

@inproceedings{dosovitskiy2021image,
  title={An image is worth 16x16 words: Transformers for image recognition at scale},
  author={Alexey Dosovitskiy and
                  Lucas Beyer and
                  Alexander Kolesnikov and
                  Dirk Weissenborn and
                  Xiaohua Zhai and
                  Thomas Unterthiner and
                  Mostafa Dehghani and
                  Matthias Minderer and
                  Georg Heigold and
                  Sylvain Gelly and
                  Jakob Uszkoreit and
                  Neil Houlsby},
  booktitle=ICLR,
  year={2021}
}

@inproceedings{he2016deep,
  title={Deep residual learning for image recognition},
  author={He, Kaiming and Zhang, Xiangyu and Ren, Shaoqing and Sun, Jian},
  booktitle=CVPR,
  pages={770--778},
  year={2016}
}

@article{zhang2025agentcpm,
  title={{AgentCPM-GUI}: Building Mobile-Use Agents with Reinforcement Fine-Tuning},
  author={Zhang, Zhong and Lu, Yaxi and Fu, Yikun and Huo, Yupeng and Yang, Shenzhi and Wu, Yesai and Si, Han and Cong, Xin and Chen, Haotian and Lin, Yankai and others},
  journal={arXiv preprint arXiv:2506.01391},
  year={2025}
}

@article{lu2025uir1,
  title={{UI-R1}: Enhancing Action Prediction of {GUI} Agents by Reinforcement Learning},
  author={Lu, Zhengxi and Chai, Yuxiang and Guo, Yaxuan and Yin, Xi and Liu, Liang and Wang, Hao and Xiong, Guanjing and Li, Hongsheng},
  journal={arXiv preprint arXiv:2503.21620},
  year={2025}
}

@article{shao2024deepseekmath,
  title={{DeepSeekMath}: Pushing the Limits of Mathematical Reasoning in Open Language Models},
  author={Shao, Zhihong and Wang, Peiyi and Zhu, Qihao and Xu, Runxin and Song, Junxiao and Bi, Xiao and Zhang, Haowei and Zhang, Mingchuan and Li, Y. K. and Wu, Y. and Guo, Daya},
  journal={arXiv preprint arXiv:2402.03300},
  year={2024}
}

@inproceedings{zhang2025saliency,
  title={Saliency-bench: A comprehensive benchmark for evaluating visual explanations},
  author={Zhang, Yifei and Song, James and Gu, Siyi and Jiang, Tianxu and Pan, Bo and Bai, Guangji and Zhao, Liang},
  booktitle=SIGKDD,
  pages={5924--5935},
  year={2025}
}

@article{gao2022large,
  title={Large-scale unsupervised semantic segmentation},
  author={Gao, Shanghua and Li, Zhong-Yu and Yang, Ming-Hsuan and Cheng, Ming-Ming and Han, Junwei and Torr, Philip},
  journal=TPAMI,
  volume={45},
  number={6},
  pages={7457--7476},
  year={2022},
  publisher={IEEE}
}

@article{selvaraju2020grad,
  title={Grad-CAM: Visual Explanations from Deep Networks via Gradient-Based Localization},
  author={Selvaraju, Ramprasaath R and Cogswell, Michael and Abhishek, Das and Ramakrishna, Vedantam and Devi, Parikh and Dhruv, Batra},
  journal=IJCV,
  volume={128},
  number={2},
  pages={336--359},
  year={2020}
}

@inproceedings{ribeiro2016should,
  title={"Why should i trust you?" Explaining the predictions of any classifier},
  author={Ribeiro, Marco Tulio and Singh, Sameer and Guestrin, Carlos},
  booktitle=SIGKDD,
  pages={1135--1144},
  year={2016}
}

@inproceedings{zhao2024gradient,
  title={Gradient-based visual explanation for transformer-based clip},
  author={Zhao, Chenyang and Wang, Kun and Zeng, Xingyu and Zhao, Rui and Chan, Antoni B},
  booktitle=ICML,
  pages={61072--61091},
  year={2024}
}

@inproceedings{zhang2025redundancy,
  title={From redundancy to relevance: Information flow in lvlms across reasoning tasks},
  author={Zhang, Xiaofeng and Quan, Yihao and Shen, Chen and Yuan, Xiaosong and Yan, Shaotian and Xie, Liang and Wang, Wenxiao and Gu, Chaochen and Tang, Hao and Ye, Jieping},
  booktitle=NAACL,
  pages={2289--2299},
  year={2025}
}

@article{xing2025large,
  title={Where do Large Vision-Language Models Look at when Answering Questions?},
  author={Xing, Xiaoying and Kuo, Chia-Wen and Fuxin, Li and Niu, Yulei and Chen, Fan and Li, Ming and Wu, Ying and Wen, Longyin and Zhu, Sijie},
  journal={arXiv preprint arXiv:2503.13891},
  year={2025}
}

@inproceedings{petsiuk2018rise,
  title={RISE: Randomized input sampling for explanation of black-box models},
  author={Petsiuk, Vitali and Das, Abir and Saenko, Kate},
  booktitle=BMVC,
  pages={151},
  year={2018}
}

@inproceedings{novello2022making,
  title={Making sense of dependence: Efficient black-box explanations using dependence measure},
  author={Novello, Paul and Fel, Thomas and Vigouroux, David},
  booktitle=NIPS,
  pages={4344--4357},
  year={2022}
}

@inproceedings{lundberg2017unified,
  title={A unified approach to interpreting model predictions},
  author={Lundberg, Scott M and Lee, Su-In},
  booktitle=NIPS,
  pages={4765--4774},
  year={2017}
}

@inproceedings{sun2023explain,
  title={Explain any concept: Segment anything meets concept-based explanation},
  author={Sun, Ao and Ma, Pingchuan and Yuan, Yuanyuan and Wang, Shuai},
  booktitle=NIPS,
  pages={21826--21840},
  year={2023}
}

@inproceedings{li2025token,
  title={Token Activation Map to Visually Explain Multimodal LLMs},
  author={Li, Yi and Wang, Hualiang and Ding, Xinpeng and Wang, Haonan and Li, Xiaomeng},
  booktitle=ICCV,
  pages={48--58},
  year={2025}
}

@inproceedings{chen2024less,
  title={Less is More: Fewer Interpretable Region via Submodular Subset Selection},
  author={Chen, Ruoyu and Zhang, Hua and Liang, Siyuan and Li, Jingzhi and Cao, Xiaochun},
  booktitle=ICLR,
  year={2024}
}

@article{chen2025less,
  title={Less is More: Efficient Black-box Attribution via Minimal Interpretable Subset Selection},
  author={Chen, Ruoyu and Liang, Siyuan and Li, Jingzhi and Liu, Shiming and Liu, Li and Zhang, Hua and Cao, Xiaochun},
  journal={arXiv preprint arXiv:2504.00470},
  year={2025}
}

@inproceedings{chen2025interpreting,
  title={Interpreting object-level foundation models via visual precision search},
  author={Chen, Ruoyu and Liang, Siyuan and Li, Jingzhi and Liu, Shiming and Li, Maosen and Huang, Zhen and Zhang, Hua and Cao, Xiaochun},
  booktitle=CVPR,
  year={2025}
}

@inproceedings{chen2025mllms,
  title={Where MLLMs Attend and What They Rely On: Explaining Autoregressive Token Generation},
  author={Chen, Ruoyu and Guo, Xiaoqing and Liu, Kangwei and Liang, Siyuan and Liu, Shiming and Zhang, Qunli and Wang, Laiyuan and Zhang, Hua and Cao, Xiaochun},
  booktitle=CVPR,
  pages={17057--17066},
  year={2026}
}

@inproceedings{simonyan2014deep,
  title={Deep inside convolutional networks: Visualising image classification models and saliency maps},
  author={Simonyan, Karen and Vedaldi, Andrea and Zisserman, Andrew},
  booktitle={ICLR 2014 Workshop},
  year={2014}
}

@article{zhang2018top,
  title={Top-down neural attention by excitation backprop},
  author={Zhang, Jianming and Bargal, Sarah Adel and Lin, Zhe and Brandt, Jonathan and Shen, Xiaohui and Sclaroff, Stan},
  journal={International Journal of Computer Vision},
  volume={126},
  number={10},
  pages={1084--1102},
  year={2018}
}

@article{gao2024going,
  title={Going beyond xai: A systematic survey for explanation-guided learning},
  author={Gao, Yuyang and Gu, Siyi and Jiang, Junji and Hong, Sungsoo Ray and Yu, Dazhou and Zhao, Liang},
  journal={ACM Computing Surveys},
  volume={56},
  number={7},
  pages={1--39},
  year={2024}
}

@article{erion2021improving,
  title={Improving performance of deep learning models with axiomatic attribution priors and expected gradients},
  author={Erion, Gabriel and Janizek, Joseph D and Sturmfels, Pascal and Lundberg, Scott M and Lee, Su-In},
  journal={Nature Machine Intelligence},
  volume={3},
  number={7},
  pages={620--631},
  year={2021}
}

@inproceedings{han2021explanation,
  title={Explanation consistency training: Facilitating consistency-based semi-supervised learning with interpretability},
  author={Han, Tao and Tu, Wei-Wei and Li, Yu-Feng},
  booktitle=AAAI,
  volume={35},
  number={9},
  pages={7639--7646},
  year={2021}
}

@inproceedings{pillai2022consistent,
  title={Consistent explanations by contrastive learning},
  author={Pillai, Vipin and Koohpayegani, Soroush Abbasi and Ouligian, Ashley and Fong, Dennis and Pirsiavash, Hamed},
  booktitle=CVPR,
  pages={10213--10222},
  year={2022}
}

@article{chen2025generalized,
  title={Generalized Semantic Contrastive Learning via Embedding Side Information for Few-Shot Object Detection},
  author={Chen, Ruoyu and Zhang, Hua and Li, Jingzhi and Liu, Li and Huang, Zhen and Cao, Xiaochun},
  journal={IEEE Transactions on Pattern Analysis and Machine Intelligence},
  volume={47},
  number={8},
  pages={6496-6514},
  year={2025}
}

@article{chen2025did,
  title={Did Models Learn Sufficiently? Attribution-Guided Training via Subset-Selected Counterfactual Augmentation},
  author={Chen, Yannan and Chen, Ruoyu and Wang, Wei and Zeng, Bin and Li, Jinke and Liu, Shiming and Zhang, Qunli and Wang, Yaowei and Cao, Xiaochun},
  journal={arXiv preprint arXiv:2511.12100},
  year={2025}
}

@inproceedings{ross2017right,
  title={Right for the right reasons: training differentiable models by constraining their explanations},
  author={Ross, Andrew Slavin and Hughes, Michael C and Doshi-Velez, Finale},
  booktitle=IJCAI,
  pages={2662--2670},
  year={2017}
}

@article{schramowski2020making,
  title={Making deep neural networks right for the right scientific reasons by interacting with their explanations},
  author={Schramowski, Patrick and Stammer, Wolfgang and Teso, Stefano and Brugger, Anna and Herbert, Franziska and Shao, Xiaoting and Luigs, Hans-Georg and Mahlein, Anne-Katrin and Kersting, Kristian},
  journal={Nature Machine Intelligence},
  volume={2},
  number={8},
  pages={476--486},
  year={2020}
}

@inproceedings{selvaraju2019taking,
  title={Taking a hint: Leveraging explanations to make vision and language models more grounded},
  author={Selvaraju, Ramprasaath R and Lee, Stefan and Shen, Yilin and Jin, Hongxia and Ghosh, Shalini and Heck, Larry and Batra, Dhruv and Parikh, Devi},
  booktitle=ICCV,
  pages={2591--2600},
  year={2019}
}

@inproceedings{zhang2023magi,
  title={Magi: Multi-annotated explanation-guided learning},
  author={Zhang, Yifei and Gu, Siyi and Gao, Yuyang and Pan, Bo and Yang, Xiaofeng and Zhao, Liang},
  booktitle={Proceedings of the IEEE/CVF International Conference on Computer Vision},
  pages={1977--1987},
  year={2023}
}

@inproceedings{rao2023studying,
  title={Studying how to efficiently and effectively guide models with explanations},
  author={Rao, Sukrut and B{\"o}hle, Moritz and Parchami-Araghi, Amin and Schiele, Bernt},
  booktitle=ICCV,
  pages={1922--1933},
  year={2023}
}

@article{zhang2024megl,
  title={MEGL: Multimodal Explanation-Guided Learning},
  author={Zhang, Yifei and Jiang, Tianxu and Pan, Bo and Wang, Jingyu and Bai, Guangji and Zhao, Liang},
  journal={arXiv preprint arXiv:2411.13053},
  year={2024}
}

@article{han2026vlaa,
  title={VLAA-GUI: Knowing When to Stop, Recover, and Search, A Modular Framework for GUI Automation},
  author={Han, Qijun and Tu, Haoqin and Wang, Zijun and Dai, Haoyue and Zhou, Yiyang and Lau, Nancy and Cardenas, Alvaro A and Xu, Yuhui and Xu, Ran and Xiong, Caiming and others},
  journal={arXiv preprint arXiv:2604.21375},
  year={2026}
}

@article{tu2026visualclaw,
  title={VisualClaw: A Real-Time, Personalized Agent for the Physical World},
  author={Tu, Haoqin and Chen, Jianwen and Wang, Zijun and Han, Siwei and Wu, Juncheng and Chen, Hardy and Ji, Haonian and Xiong, Kaiwen and Liu, Jiaqi and Xia, Peng and others},
  journal={arXiv preprint arXiv:2606.16295},
  year={2026}
}

@inproceedings{stojanov2022learning,
  title={Learning dense object descriptors from multiple views for low-shot category generalization},
  author={Stojanov, Stefan and Thai, Anh and Huang, Zixuan and Rehg, James},
  booktitle=NIPS,
  pages={12566--12580},
  year={2022}
}

@inproceedings{xiao2020noise,
  title={Noise or signal: The role of image backgrounds in object recognition},
  author={Xiao, Kai and Engstrom, Logan and Ilyas, Andrew and Madry, Aleksander},
  booktitle=ICLR,
  year={2021}
}
\bibliographystyle{tmlr}

\appendix
\setcounter{figure}{0}

\renewcommand{\thefigure}{\thesection\arabic{figure}}

\section{More Details for GUI-Agent Experiments}

\subsection{Dataset Format}

Each sample is stored as a JSON object containing an \texttt{id}, a screenshot reference (\texttt{image}), and a multi-turn \texttt{conversations} list in a \texttt{system/user/assistant} format.
The assistant output follows a strict action schema and includes a \texttt{thought} field (thinking) and an executable action such as a click \texttt{POINT} (decision).
In addition, each sample provides a human-annotated \texttt{bounding\_box} for the target UI element, which serves as a human prior for alignment and evaluation.
All GUI tasks are collected and executed in Chinese, and the paper presents an English-translated version of the prompts for clarity.

\begin{jsoncode}
{
  "id": "0",
  "image": {
    "<image_00>": "img/screenshot_0.jpg"
  },
  "conversations": [
    {
      "role": "system",
      "content": "# Role\nYou are an agent familiar with Android touch-based GUI operations. Given a user's request, analyze the GUI elements and layout on the current screen and produce the next action.\n\n# Task\nGiven the current screenshot, output the next operation to accomplish the user request.\n\n# Rule\n- Output in compact JSON format.\n- The action must follow the Schema constraints.\n\n# Schema\n{\"type\":\"object\",\"description\":\"Execute an action and decide the task status\",\"additionalProperties\":false,\"optional\":[\"thought\"],\"properties\": {\"thought\":{\"type\":\"string\",\"description\":\"The agent's reasoning\"}, \"POINT\":{\"$ref\":\"#/$defs/Location\",\"description\":\"Click a specific position on the screen\"},\"to\":{\"description\":\"Movement / gesture parameters\", \"oneOf\":[{\"enum\":[\"up\",\"down\",\"left\",\"right\"],\"description\":\"Swipe from the current point (POINT) in one of four directions\"},{\"$ref\":\"#/$defs/Location\",\"description\":\"Move to a specific location\"}]},\"duration\":{\"type\":\"integer\",\"description\":\"Execution or wait time in milliseconds\",\"minimum\":0,\"default\":200},\"PRESS\":{\"type\":\"string\",\"description\":\"Trigger a special key\",\"enum\":[\"HOME\",\"BACK\",\"ENTER\"]},\"TYPE\":{\"type\":\"string\",\"description\":\"Input text\"},\"STATUS\":{\"type\":\"string\",\"description\":\"Task status: satisfied (no action needed), impossible, interrupt, need_feedback\", \"enum\":[\"continue\",\"finish\",\"satisfied\",\"impossible\",\"interrupt\",\"need_feedback\"], \"default\":\"continue\"}},\"\$defs\":{\"Location\":{\"type\":\"array\",\"description\":\"Coordinates are relative to the top-left corner and scaled to [0,1000]; the first entry is x and the second is y\",\"items\":{\"type\":\"integer\",\"minimum\":0,\"maximum\":1000},\"minItems\":2,\"maxItems\":2}}}"
    },
    {
      "role": "user",
      "content": "<Question>Search bilibili.com, then search for 'Qianting Weiwei Mi' on the website and open the uploader list. Add the page to bookmarks and verify it appears in the bookmark manager.</Question>\nCurrent screenshot: <image_00>"
    },
    {
      "role": "assistant",
      "content": "{\"thought\":\"Locate and tap the browser icon on the home screen to open the browser.\",\"POINT\":[591,915]}"
    }
  ],
  "bounding_box": [706,2438,858,2590]
}
\end{jsoncode}

\subsection{Attributing Thinking and Decision}

Unlike approaches that attribute only the final action, we treat the mobile agent's \emph{reasoning process} (thinking) and \emph{executed action} (decision) as one supervised autoregressive sequence. Given model parameters $\theta$ and input $X$, EAGLE jointly attributes the visual evidence associated with generating $\hat{Z}$ (the reasoning sequence) and $\hat{A}$ (the final action decision):
\begin{equation}
\pi_{\mathrm{gui}} = \operatorname{EAGLE}(f_\theta, X, \{\hat{Z}, \hat{A}\}),
\end{equation}
Here, $\pi_{\mathrm{gui}}$ is the attribution ranking over screen regions for the complete thinking--action sequence. EAGLE scores a retained subset and its complement with $\mathcal{F}^{\mathrm{gui}}_\theta$, defined in Section~\ref{prior:loss}, by summing the target-token probabilities at the supervised thinking--action positions. The resulting ranking is mapped to screen coordinates and compared with the target UI bounding box.

\subsection{Training Procedure}

To control computational overhead, the attribution alignment loss is computed periodically rather than at every update step. At these steps, EAGLE produces an ordered set of at most $k$ screen regions, and each region is classified as prior-consistent or off-prior using the overlap threshold $\tau$. The discrete ranking and overlap gates are detached; gradients are computed only through $\mathcal{F}^{\mathrm{gui}}_\theta$ on the fixed masked screenshots. Algorithm~\ref{alg:gui_attr_train} formally defines the previously unspecified attribution penalty.

\begin{algorithm}[!htb]
    \caption{Attribution-guided prior alignment for GUI agents}
    \label{alg:gui_attr_train}

    \KwIn{Training sample $(X,I,Z,A,B)$; GUI set function $\mathcal{F}^{\mathrm{gui}}_\theta$; interval $K$; loss weights $\lambda_1,\lambda_2$; overlap threshold $\tau$; maximum attribution length $k$, where $k=2$.}
    \KwOut{Trained model parameters $\theta$.}

    Initialize model parameters $\theta$\;

    \For{$t=1$ \KwTo $T_{\max}$}{
        Predict thinking $\hat{Z}$ and action $\hat{A}$ under current parameters $\theta$\;
        Compute supervised loss $\mathcal{L}_{\mathrm{CE}}$ \Comment*[r]{SFT cross-entropy loss}

        \uIf{$t \bmod K == 0$}{
            $\pi=(v_{\pi_1},\ldots,v_{\pi_k}) \gets \operatorname{EAGLE}(\mathcal{F}^{\mathrm{gui}}_\theta,I,\{Z,A\})$\;
            Detach the discrete ranking, top-$k$ selection, stopping decision, and overlap indicators\;
            $\mathcal{L}_{\mathrm{deviation}}\gets0$, $\mathcal{L}_{\mathrm{redundancy}}\gets0$\;

            \uIf{$\phi(v_{\pi_1},B)<\tau$}{
                $\mathcal{L}_{\mathrm{deviation}}\gets\mathcal{F}^{\mathrm{gui}}_\theta(\{v_{\pi_1}\})$\;
            }
            \For{$r=2$ \KwTo $k$}{
                \uIf{$\phi(v_{\pi_r},B)<\tau$}{
                    $\Delta_r\gets\mathcal{F}^{\mathrm{gui}}_\theta(S_r^\pi)-\operatorname{sg}[\mathcal{F}^{\mathrm{gui}}_\theta(S_{r-1}^\pi)]$\;
                    $\mathcal{L}_{\mathrm{redundancy}}\gets\mathcal{L}_{\mathrm{redundancy}}+\mathsf{ReLU}(\Delta_r)$\;
                }
            }
            $\mathcal{L}_{\mathrm{attr}}\gets\lambda_1\mathcal{L}_{\mathrm{deviation}}+\lambda_2\mathcal{L}_{\mathrm{redundancy}}$\;

            $\mathcal{L}_{\mathrm{total}} \gets
            \mathcal{L}_{\mathrm{CE}} + \mathcal{L}_{\mathrm{attr}}$\;
        }
        \Else{
            $\mathcal{L}_{\mathrm{total}} \gets \mathcal{L}_{\mathrm{CE}}$\;
        }
        Update model parameters $\theta$ using $\nabla_{\theta}\mathcal{L}_{\mathrm{total}}$\;
    }
    \Return $\theta$\;
\end{algorithm}

\FloatBarrier
\subsection{Additional GUI Training Controls}

\textbf{Box-Guided Augmentation SFT.} Inspired by prior foreground--background intervention and background-randomization controls~\citep{xiao2020noise,stojanov2022learning}, this control receives the same target UI bounding box as our method but does not compute an attribution. With probability $0.5$, it uses the original screenshot. Otherwise, the target box is expanded by $10\%$ in each spatial direction and the pixels outside the expanded box are randomly either replaced by the image-preprocessing mean or Gaussian blurred, with equal probability. The screenshot size and click coordinates are unchanged. The model optimizes only the ordinary token-level SFT cross-entropy on the target thinking--action sequence. At validation and test time, it receives the complete, unmodified screenshot and no bounding box.

\textbf{Task-Reward GRPO.} Following GRPO~\citep{shao2024deepseekmath} and its rule-reward use for GUI action prediction~\citep{lu2025uir1,zhang2025agentcpm}, this control starts from the SFT checkpoint and performs Group Relative Policy Optimization using only verifiable output and task rewards. It does not use EAGLE, Point Game, or any attribution score. For a generated response $\hat{y}$, the reward is
\begin{equation}
R(\hat{y})=
\begin{cases}
-1, & \text{invalid JSON/schema or empty thinking},\\
-0.5, & \text{incorrect action type},\\
0.2+0.2+0.6g, & \text{valid format and correct action type},
\end{cases}
\end{equation}
where the first two positive terms reward schema validity and the correct action type. A correct non-\texttt{POINT} action uses $g=1$. For a predicted click $p$ and target box $B$, let $d(p,B)$ be the Euclidean distance to the box in the $[0,1000]^2$ action-coordinate system. The grounding reward is
\begin{equation}
g=
\begin{cases}
1, & p\in B,\\
-\min\!\left\{\dfrac{d(p,B)/(1000\sqrt{2})}{0.25},1\right\}, & p\notin B.
\end{cases}
\end{equation}
Thus, an in-box click receives total reward $1$, whereas a valid off-box click receives a distance-shaped reward ranging from approximately $0.4$ to $-0.2$. Each prompt produces four sampled responses. Their group-relative advantages are
\begin{equation}
A_i=\frac{R_i-\operatorname{mean}(R)}{\operatorname{std}(R)+10^{-4}}.
\end{equation}
When all four rewards are identical, all advantages are zero. We use a KL coefficient of $\beta=0.01$ relative to the frozen SFT reference policy.

\subsection{Evaluation Metrics}\label{prior:gui_evaluation_metrics}

\begin{figure*}[!htbp]
    \centering
    \includegraphics[width=\textwidth]{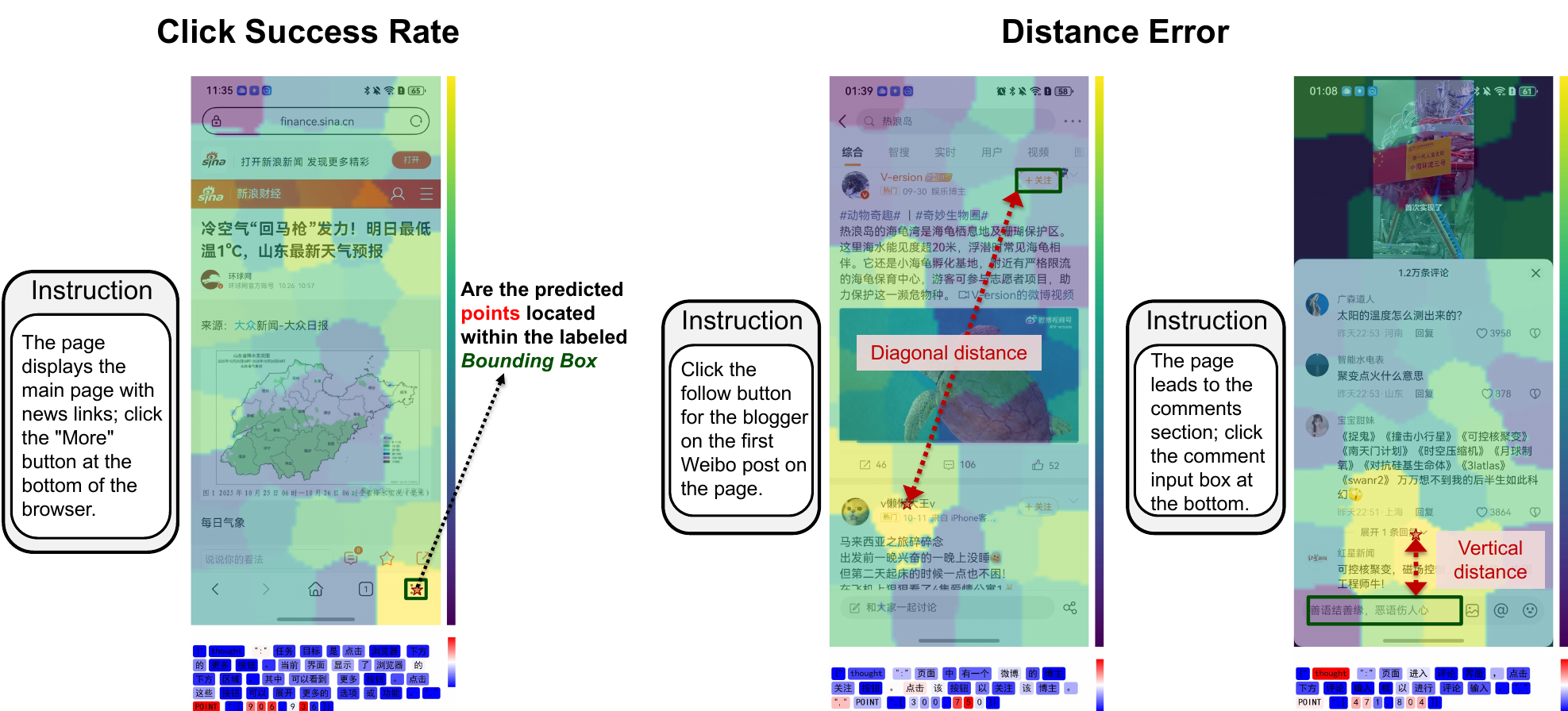}
    \caption{Illustration of evaluation metrics for the GUI agent clicking task. \emph{Click Success Rate} counts a prediction as successful if the predicted click point falls inside the labeled target UI element bounding box. \emph{Distance Error} measures the distance between the predicted click point and the target location (e.g., to the bounding box or target point), with examples showing diagonal and vertical distances.}
    \label{prior_align:evaluation_metric_gui}
\end{figure*}

Two metrics are used to evaluate the GUI agent clicking task: click success rate and distance error.
The \textbf{click success rate} measures whether the predicted click point falls inside the human-annotated target bounding box.
The \textbf{distance error} quantifies how far the predicted point is from the ground-truth target region: it is set to $0$ if the predicted point lies inside the bounding box; otherwise, it is computed as the minimum Euclidean distance from the point to the bounding box boundary (i.e., the closest point on the box). Figure~\ref{prior_align:evaluation_metric_gui} illustrates these metrics with representative examples.

\FloatBarrier
\subsection{Computational Cost and Reproducibility}

Subset-attribution alignment adds training-time computation because it evaluates multiple masked inputs during the attribution search. We reduce this cost by applying alignment once every 10 optimization steps for classification and once every 5 steps for the GUI agent. For classification, the search is restricted to correctly predicted samples with confidence above $0.75$, considers at most 10 of 50 regions, and early-stops when selected-subset confidence exceeds $0.8$; the GUI experiment considers at most two regions. The greedy search, top-$k$ selection, stopping decision, and overlap gates are detached, so the full discrete search does not retain a gradient graph. Gradients are stored only for the fixed masked inputs used by the alignment loss. The additional cost is confined to training: the trained model has the same architecture and inference procedure as its fine-tuned counterpart.

\begin{table}[!htbp]
    \caption{Representative classification wall-clock time per epoch. Each fine-tuning/ours pair was measured under the same setup; absolute time is hardware-dependent.}
    \begin{center}
        \resizebox{0.6\textwidth}{!}{
        \begin{tabular}{ccrrr}
            \toprule
            \textbf{Dataset} & \textbf{Model} & \textbf{Fine-tuning} & \textbf{Ours} & \textbf{Ratio} \\
            \midrule
            \multirow{3}{*}{Saliency-Bench}
            & CLIP       & 43 s     & 7 m 50 s & $10.93\times$ \\
            & ViT (base) & 37 s     & 2 m 42 s & $4.38\times$ \\
            & ResNet-101 & 35 s     & 1 m 04 s & $1.83\times$ \\
            \midrule
            \multirow{3}{*}{ImageNet-S}
            & CLIP       & 2 m 18 s & 8 m 34 s & $3.72\times$ \\
            & ViT (base) & 1 m 04 s & 2 m 54 s & $2.72\times$ \\
            & ResNet-101 & 23 s     & 2 m 39 s & $6.91\times$ \\
            \bottomrule
        \end{tabular}}
    \end{center}
    \label{prior:training_cost}
\end{table}

The wall-clock overhead in Table~\ref{prior:training_cost} is model- and dataset-dependent rather than negligible. Upon publication, we will release the training and evaluation code, complete configuration files, environment specification, random seeds, and dataset split identifiers. We will also release the GUI data-construction and annotation protocol and all data or metadata that can be redistributed subject to screenshot privacy and licensing constraints.

\section{Limitations and Future Work}

\textbf{Limitations.} The method depends on the quality and coverage of human-prior annotations. A mask or bounding box is weak, potentially incomplete supervision rather than causal ground truth: it may omit legitimate contextual evidence, reflect annotator assumptions, or fail to transfer across interface conventions, languages, and accessibility settings. Coarse or noisy priors may weaken or bias the training signal. Finally, each GUI method uses one training run on the 936-task dataset, so training variance and statistical significance remain uncharacterized.

\textbf{Additional supervision and scalability.} Our method requires spatial priors in addition to task labels. In the classification experiments, we reuse masks already provided by Saliency-Bench and ImageNet-S, but obtaining comparable masks in new domains can be costly. The GUI study additionally requires target-element bounding boxes, whose collection and quality control increase the supervision burden beyond standard SFT. Consequently, comparisons with no-prior fine-tuning do not match annotation budgets and should be interpreted as measuring the benefit of added prior supervision, while the matched-mask and matched-box controls provide fairer comparisons under the same spatial annotations. This annotation--performance trade-off limits scalability when reliable priors are unavailable. Weakly supervised or automatically generated priors may reduce cost, but can introduce noise or systematic bias and therefore require independent validation.

\textbf{Future work} should evaluate sensitivity to noisy or conflicting priors, test across diverse interfaces and accessibility configurations, and obtain repeated GUI training runs. Scalable prior acquisition could use weak/self-supervised cues, pseudo-labels, or multi-model consensus, but such generated priors would require independent auditing. A further operator-matched ablation could aggregate gradient- or CAM-based heatmaps into the same region space and hold the masked-input set function fixed; the aggregation and ordering choices would need to be controlled explicitly. Attribution-based objectives could also be studied in reinforcement learning while retaining human confirmation and restricted permissions for consequential actions.

\end{document}